\documentclass[sigconf]{acmart}
\usepackage[nomargin,inline,final]{fixme}
\usepackage{url}            
\usepackage{booktabs}       
\usepackage{amsfonts}       
\usepackage{nicefrac}       
\usepackage{microtype}      
\usepackage{xcolor}         
\usepackage{graphicx, subcaption}
\usepackage{doi}
\usepackage{gensymb}
\usepackage{algorithm}
\usepackage{algpseudocode}
\usepackage{amsmath}
\usepackage{soul}
\usepackage{siunitx}
\usepackage{enumitem}
\usepackage{multirow}
\usepackage{comment}
\usepackage{pifont}
\usepackage{makecell}
\newcommand{\xmark}{\ding{55}}%
\usepackage{animate}

\newcolumntype{I}{!{\vrule width 1.2pt}}
\newlength\savedwidth

\newlength\savewidth

\AtBeginDocument{%
  \providecommand\BibTeX{{%
    \normalfont B\kern-0.5em{\scshape i\kern-0.25em b}\kern-0.8em\TeX}}}

\begin{document}

\title{CRA5: Extreme Compression of ERA5 for Portable Global Climate and Weather Research via an Efficient Variational Transformer}

\author{Tao Han$^\ddagger{}^*$, Zhenghao Chen$^\dag$, Song Guo$^*$, Wanghan Xu$^\ddagger$, Lei Bai$^\ddagger$\\
$^\ddagger$ Shanghai Artificial Intelligence Laboratory, $^*$The Hong Kong University of Science and Technology, $^\dag$ The University of Sydney}
\thanks{Corresponding Author: Lei Bai, \href{mailto:baisanshi@gmail.com}{baisanshi@gmail.com}}

\settopmatter{printacmref=false} 
\renewcommand\footnotetextcopyrightpermission[1]{} 
\pagestyle{plain} 

\begin{abstract}
The advent of data-driven weather forecasting models, which learn from hundreds of terabytes (TB) of reanalysis data, has significantly advanced forecasting capabilities. However, the substantial costs associated with data storage and transmission present a major challenge for data providers and users, affecting resource-constrained researchers and limiting their accessibility to participate in AI-based meteorological research. To mitigate this issue, we introduce an efficient neural codec, the Variational Autoencoder Transformer (\emph{VAEformer}), for extreme compression of climate data to significantly reduce data storage cost, making AI-based meteorological research portable to researchers. Our approach diverges from recent complex neural codecs by utilizing a low-complexity Auto-Encoder transformer. This encoder produces a quantized latent representation through variance inference, which reparameterizes the latent space as a Gaussian distribution. This method improves the estimation of distributions for cross-entropy coding. Extensive experiments demonstrate that our VAEformer outperforms existing state-of-the-art compression methods in the context of climate data. By applying our VAEformer, we compressed the most popular ERA5 climate dataset (226 TB) into a new dataset, CRA5 (0.7 TB). This translates to a compression ratio of over 300 while retaining the dataset’s utility for accurate scientific analysis. Further, downstream experiments show that global weather forecasting models trained on the compact CRA5 dataset achieve forecasting accuracy comparable to the model trained on the original dataset. Code, the CRA5 dataset, and the pre-trained model are available at \textcolor{blue}{\url{https://github.com/taohan10200/CRA5}}.

\end{abstract}






\begin{teaserfigure}
\centering
 \includegraphics[width=\textwidth]{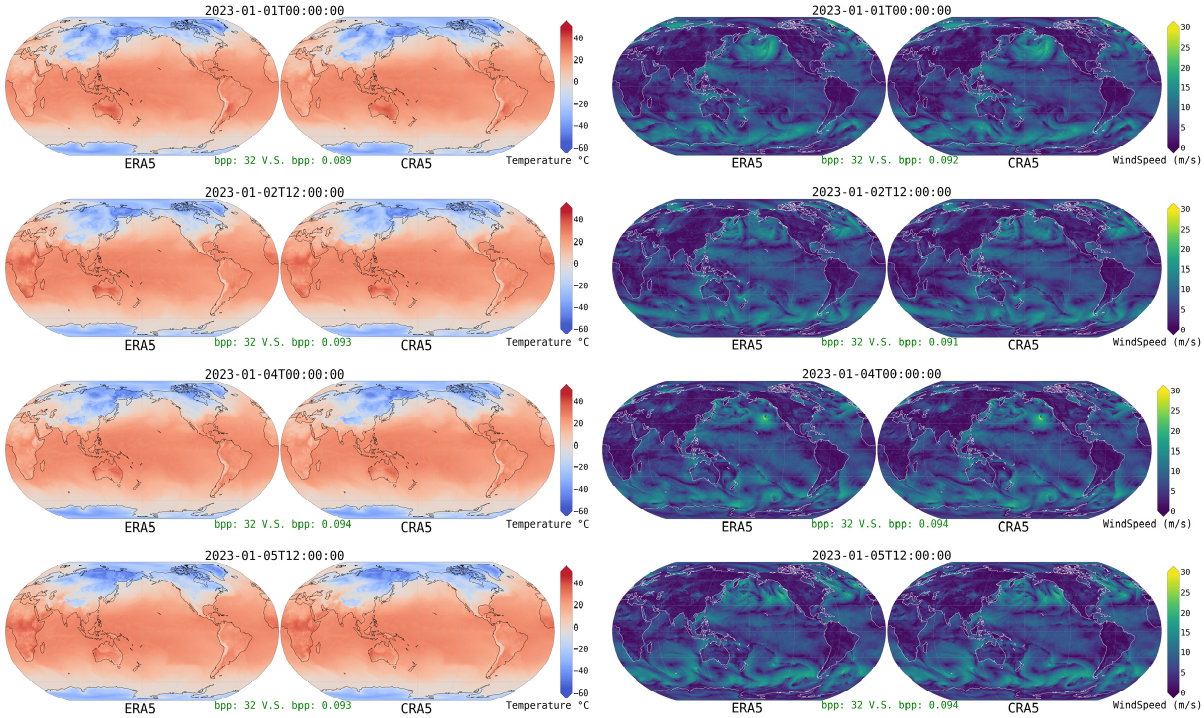}
\caption{This demonstration showcases the sample data of the original ERA5 (\textcolor{red}{226TB}) dataset and  CRA5 (\textcolor{red}{0.7TB}) dataset compressed by the proposed \textit{VAEformer}. It is recommended that you use Adobe Acrobat Reader for optimal viewing.}
\end{teaserfigure}

\maketitle

\section{Introduction}
Climate change has aroused awareness of understanding and predicting the operational earth system at an ever-rising speed. To track such climate change better and provide more accurate weather prediction for society, many super-computing centres in the world are capturing the atmospheric data from different devices, such as weather stations, radiosondes, and satellites~\cite{schultz2021can}, to run the operational weather and climate simulations several times on a daily basis. Therefore, meteorological data, such as observational data and derivative products, grow at the petabyte scale one day.

On the other hand, high costs of data storage and transmission stand as the main obstacles to progress in climate and weather research, leading to several related challenges:
\textbf{1) Climate data storage expenditure}: 
Statistics from the European Centre for Medium-Range Weather Forecasts (ECMWF) show that its archive grows by about 287 terabytes (TB) on an average day~\footnote{\url{https://www.ecmwf.int/en/computing/our-facilities/data-handling-system}}, while its data production is 
and will be predicted to quadruple within the next decade~\cite{klower2021compressing}. 
\textbf{2) Distribution difficulty of weather service}: 
Weather forecasting groups are facing an obstacle to distributing their products because handling and transmitting terabyte-scale data daily can overload networks and become costly (\textit{e.g.,} 215 TB of data in ECMWF is retrieved daily by end users worldwide.).
\textbf{3) High-threshold data-driven weather and climate research}: 
An increasing trend in Numerical Weather Prediction (NWP) involves deep learning techniques \cite{rasp2020weatherbench,hu2023swinvrnn,pathak2022fourcastnet,bi2022pangu,lam2022graphcast,chen2023fengwu,huang2023compressing,price2023gencast}, benefiting from the extensive dataset (\textit{e.g.,} ERA5~\cite{hersbach2020era5}) with over $1,000$ TB of data, which is impractical for individual researchers to manage and store on personal devices. 

To address the first challenge, we present an efficient transformer-based compression framework tailored to this work's practical lossy climate data compression task. Neural Image Compression (NIC) methods directly adopt Auto-Encoder (AE) style network~\cite{kingma2013auto}. However, we propose a dual Variantional-Auto-Encoder (VAE) style network, which contains two VAEs for generating quantized latent features and hyper-prior information, respectively. Unlike previous NIC methods, which transform the image data into a non-regulation representation, we encode the weather data into a latent with Gaussian distribution feature using a variational inference process. Therefore, the encoder turns to predict the distribution of weather data $X$, where latent is sampled from the learned distribution $\mathcal{N}(\mu_x, \sigma_x)$ instead of simply transforming an input as a direct representation as in most nowadays NICs. It ensures our produced latent representation $\hat{y}$ has good distribution, allowing the entropy model to estimate its distribution more precisely. 
As for the entropy model, we adopt a similar network architecture but use a variational inference process to simulate the quantization process. It is inspired by common NIC paradigm~\cite{balle2018variational,minnen2018joint,cheng2020learned,qian2022entroformer, chen2022exploiting, chen2022lsvc, hu2020improving, liu2023icmh}, which embeds a VAE as a parametrized entropy model.

To ensure a practical encoding time, we propose an efficient transformer architecture for the deploying of two transformer-based VAEs, where a large portion of earth-specific window-attention blocks decreases the complexity from $O(N^2)$ to $O(N)$ and several global attention blocks are kept to capture the global circulation of the earth-atmosphere system~\cite{sellers1969global,saha2008earth,hill2004architecture}. In this way, our codec can be more time-effective when encoding oversized weather data.
Moreover, to enable a practical optimization of our VAEformer,  we decompose the training procedure into the pre-trained reconstruction part and the entropy coding model fine-tuning part, which stabilizes the training process and improves compression performance with a large pre-trained reconstruction model. 
Comprehensive experiments conducted on the ERA5 data compression show our proposed framework outperforms the conventional NIC methods~\cite{balle2018variational,minnen2018joint,cheng2020learned, zou2022devil,he2022elic,liu2023learned,xie2021enhanced}.

To address the second challenge, we propose a new paradigm for weather service delivery. 
Specifically, data providers compress weather data into highly compact quantized representations using our VAEformer before broadcasting, enabling efficient data distribution with minimal network bandwidth requirements. 
In this way, we can alleviate the need for users to allocate substantial storage space for storing the raw over-size data. 
Instead, users only need to access the compressed representations from the cloud side and execute the decoding procedure on their side to retrieve and visualize the desired information rapidly. 
As a proof-of-concept, we conduct a large compression experiment to compact the most representative and used dataset ERA5 with~\textcolor{red}{226 TB} into a Compressed-ERA5 (CRA5) dataset with only \textcolor{red}{0.7 TB}. By utilizing VAEformer, our experiment demonstrates a compression ratio of more than \textbf{$300\times$} while simultaneously preserving the numerical accuracy and important information (\emph{e.g.}, extreme values) for climate research.

To address the third challenge, we utilize our produced CRA5 dataset to train a data-driven global numerical weather forecast model. A comparative analysis showcases that the performance of the CRA5-trained model is evaluated against models trained on the original ERA5, ECMWF’s HRES forecasts~\cite{81380}, and Pangu-Weather~\cite{bi2022pangu}. The findings indicate that the model trained on CRA5 achieves forecast skills comparable to that of the model trained on ERA5. Notably, leveraging CRA5 facilitated the development of competitive models with both physics-based and AI-based models. These results reveal the potential of climate data compression to alleviate data burdens, enabling more researchers to engage in meteorological studies more conveniently. This outcome, in turn, would contribute to humanity's ability to effectively manage climate change, natural disasters, and related challenges.
The contributions of this paper are summarized as follows: 

\vspace{-1mm}
\begin{itemize}
    \item We introduce a novel and efficient neural codec, VAEformer, for weather and climate data compression, in which VAE performs variational inference to regularize the latent distribution instead of simply transforming the data into an alternative representation, resulting in a more predictable latent feature and supporting better cross-entropy coding.

    \item We represent a pioneering effort to make large-scale climate data compression a practice, where we extremely compress the most popular weather and climate research dataset ERA of 226TB to a new compact dataset CRA of 0.7TB while maintaining high numerical accuracy and the ability to recover extreme values, significantly reducing the costs associated with climate data storage and transmission. 

    \item  We develop a global numerical weather prediction model using the compressed CRA5, demonstrating that the compressed data could train models with equivalent accuracy and competitive forecasting skills compared with the best physical-based numerical model. This promising result would facilitate the convenient engagement of researchers in weather and climate studies.
    
\end{itemize}

\vspace{-2mm}
\section{Related work}

\subsection{Neural Image Compression}
Our work draws significant inspiration from the recent advances in practical NIC methods, which have shown remarkable success and surpassed traditional hand-crafted image codecs~\cite{wallace1992jpeg, taubman2002jpeg2000, bellard2015bpg}. Such NIC methods typically employ a dual AE structure. For instance, Ballé \emph{et al.} introduced a \emph{hyper-prior-based} approach, employing one AE to transform the input image into a quantized latent representation in a lossy manner, followed by a second AE that functions as an embedded entropy model for lossless encoding~\cite{balle2016end, balle2018variational, chen2022exploiting}.
Building on this, Minnen~\emph{et al.} enhanced the model by incorporating context information from latent features, proposing an \emph{auto-regressive-based} method \cite{minnen2018joint}. Recently, Cheng~\emph{et al.} have integrated a Gaussian Mixture Model and a simple attention mechanism into this dual AE-style NIC paradigm \cite{cheng2020learned}.
Moreover, integrating a more powerful transformer~\cite{vaswani2017attention} module has further propelled NIC advancements~\cite{vaswani2017attention}. For example, the Entroformer~\cite{qian2022entroformer} utilizes a Vision Transformer (ViT) Encoder as its hyper-prior model and a ViT Decoder as the context model, while to mitigate the complexity inherent in ViT, Contextformer~\cite{koyuncu2022contextformer} employs a sliding-window mechanism in its entropy models 
In addition to purely ViT-based codecs, other research has explored alternative architectures. For instance, some have adopted Swin Transformer~\cite{hu2023swinvrnn} style codecs~\cite{zhu2021transformer, zou2022devil}, while others have developed efficient parallel transformer-CNN hybrid models~\cite{liu2023learned, chen2023stxct}.

Even though existing NIC methods can intuitively transform raw images into latent representations of various distributions, predicting these distributions remains challenging. This unpredictability stems from the unconstrained transformations, which poses difficulties for subsequent entropy coding processes. To address this issue, our approach leverages a variational inference process to generate latent features, which ensures a more predictable and robust distribution, thereby simplifying the task for the entropy model in estimating the distribution accurately.

\subsection{Weather Data Compression}
For the conventional weather data compression, the Copernicus Atmospheric Monitoring Service (CAMS)~\cite{inness2019cams} adopts a linear quantization to compress the data, which is the widely used GRIB2~\cite{dey2007guide} format for storing atmospherical data. Klöwer~\emph{et al.}~\cite{klower2021compressing} define the bitwise real information content from information theory for the CAMS, where most variables contain fewer than 7 bits of real information per value and are highly compressible. All CAMS data are 17× compressed relative to 64-bit floats. COIN++~\cite{dupont2022coin++} proposes a neural compression framework that handles a wide range of data modalities, including weather data, which stores modulations applied to a meta-learned base network as a compressed code for each data. Similarly, Huang~\emph{et al.}~\cite{huang2022compressing} propose a coordinate-based neural network to overfit the data, and the resulting parameters are taken as a compact representation of the multidimensional weather and climate data. These two neural network-based methods for weather data both overfit a piece of data and thus can not be directly applied to unseen data.

\vspace{-2mm}
\subsection{Data-driven Weather Forecasting}
From 2022, data-drive numerical weather forecast models ~\cite{bi2022pangu,lam2022graphcast,chen2023fengwu,kochkov2023neural,kurth2023fourcastnet} attract significant attention in both AI and atmospheric science communities. They gradually show the potential to surpass the state-of-the-art physical-based NWP model~\cite{charney1950numerical,lynch2008origins} in terms of forecast skill and operational efficiency. 
FourCastNet~\cite{kurth2023fourcastnet} is the first global $0.25^{\circ}$ AI-based weather model to show competitive performance versus the physical-based model on some variables. Pangu-Weather~\cite{bi2022pangu} also runs on $0.25^{\circ}$ and 7-day lead times and is shown to outperform HRES on many variables. GraphCast~\cite{lam2022graphcast} is a GNN-based model operated on $0.25^{\circ}$, 37 vertical levels, and 10-day lead times, and has proven to outperform the physical model comprehensively. FengWu~\cite{chen2023fengwu} further improves the forecast skills compared with GraphCast, emphasising longer lead-time RMSE scores using a novel replay buffer technique. Recently, FengWu-GHR~\cite{han2024fengwu} lifts the resolution to $0.09^{\circ}$, which is the first AI-based model to align with the most advanced Integrated Forecasting System~\cite{81380}, showing the ever-increasing data demands in this field.

\section{Methodolgy}

\begin{figure*}

  \centering      \includegraphics[width=0.96\textwidth]{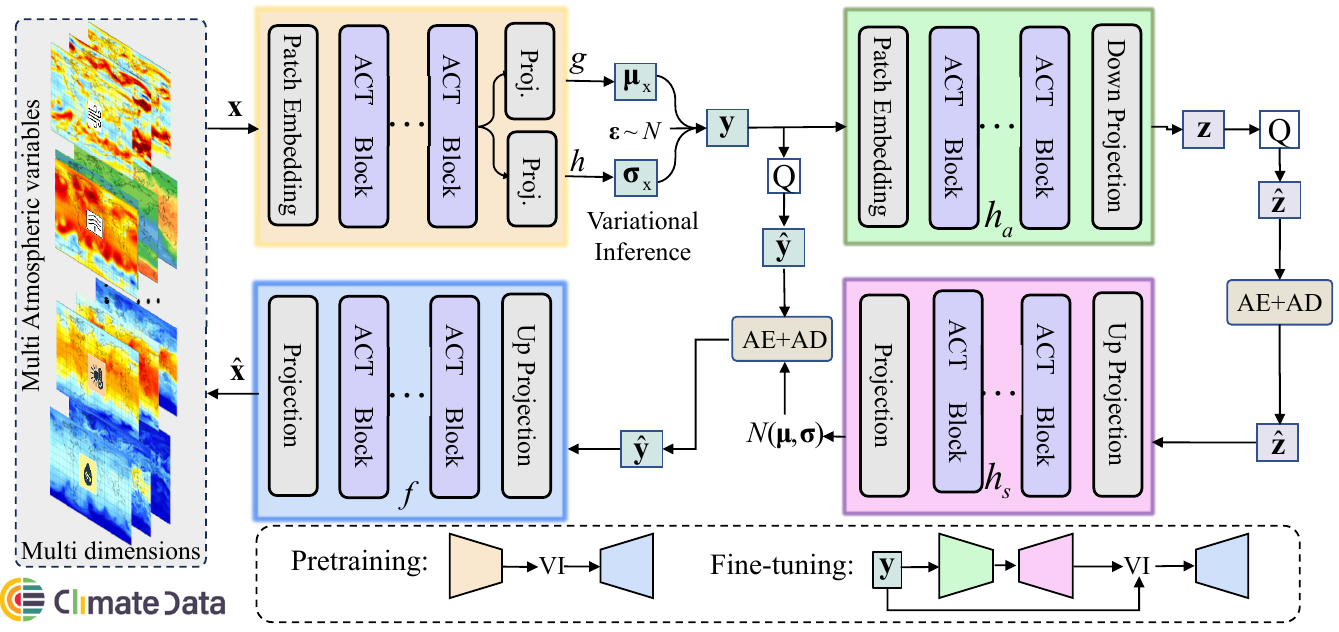}
    \caption{Overview of the proposed efficient VAEformer with our Atmospheric Circulation Transformer Block (ACT) blocks (details in Fig.~\ref{fig:ACT_block}) for the weather data compression. 
    It compresses data in both pretraining and fine-tuning:
    1) Pretraining: a VAE-style transformer encoder generates the compressed latent representation $y$, and a transformer-based decoder restores it to the reconstructed data.
    2) Fine-tuning: we train another encoder and decoder to predict the mean and scale hyperpriors for the Arithmetic Encoder and Decoder (AE+AD) process, which further losslessly compresses the data via entropy coding.
    }
    \label{fig:net}

\end{figure*}

\subsection{Overview of Compression Framework }
\label{sec:compreesor}
\textbf{Preliminaries.}
Let $x \in \mathbb{R}^{C \times H \times W}$ represent a high-dimension atmosphere data (\emph{i.e.,} 3D weather state tensors), where $C$ denotes the number of atmosphere variables, $H$ and $W$ are the latitude height and the longitude width and. 
For example, in ERA5 reanalysis dataset, we may have $C=159, W=721$, and $H=1440$. 

This work aims to compress $x$ using the proposed framework shown in Fig.~\ref{fig:net}.
We denote our latent transformer encoders with learnable parameters as two mapping functions $\boldsymbol{g}(.)$ and $\boldsymbol{h}(.)$, which produce mean and variance, followed by using the sample function  
 $\mathcal{S}$ introduced in Sec.~\ref{sec:pretain} and quantization process $Q=\lfloor\cdot \rceil$ to produce the latent representation $\hat{y}$. 
Then we denote the hyperprior transformer encoder as one mapping function $\boldsymbol{h}_a(.)$, which produces the hyperprior feature and quantizes it as $\hat{z}$. The process for generating $\hat{y}$ and $\hat{z}$ are represented as follows,

\begin{equation} \label{eq:formualtion}
\hat{y} = \boldsymbol{Q}\left(\mathcal{S} \left(\boldsymbol{g}(x), \boldsymbol{h}(x)\right)\right), \hat{z} = \boldsymbol{Q} \left(\boldsymbol{h}_a(y)\right),
\end{equation}

\textbf{Hyper-prior-based dual-VAE architecture} 
The proposed VAEformer adopts the similar architecture of hyperperior-based entropy models~\cite{balle2018variational,minnen2018joint,cheng2020learned,qian2022entroformer}, which utilizes quantized latent hyperprior feature $\hat{z}$ as prior information to match  $p_{\hat{\boldsymbol{y}}}$  more closely to the marginal for a particular weather tensor.
Fig.~\ref{fig:net} depicts a high-level overview of the transformer-based compression method, comprising two encoding-decoding processes. 
We first adopt a Transformer-based VAE (the left component of Fig.~\ref{fig:net}), which aims to learn the distribution of the original dataset. In this way, we can sample the latent representation $\hat{\boldsymbol{y}}$ from the learned distribution $\mathcal{N}(\boldsymbol{\mu_x}, \boldsymbol{\sigma_{x}})$ instead of using the straightforward medium latent in previous methods~\cite{balle2016end,balle2018variational,minnen2018joint,cheng2020learned,qian2022entroformer, leecontext,minnen2020channel}. We then use another Transformer-based VAE as our entropy model (The right component of Fig.~\ref{fig:net}), whose responsibility is to learn a probabilistic model over quantized latent $\hat{z}$ for conditional entropy coding. Specifically, we use a transformer-based hyper-prior decoder to produce the mean $\boldsymbol{\mu}$ and scale $\boldsymbol{\sigma}$ parameters for a conditional Gaussian entropy model to perform a lossless compression over the quantized $\hat{y}$. 

Where $\boldsymbol{\theta}_{{\boldsymbol{h}}_s}$ and $\boldsymbol{\theta_{ep}}$ respectively represent the parameters in the hyperprior decoder and the entropy parameters network. For the typical NIC task, the compressed latent and the compressed hyperprior are parts of the ultimately generated bitstream. Therefore, for this task, we need to optimize the trade-off between the estimated coding length of the bitstream and the quality of the reconstruction. Hence,  the rate-distortion optimization problem is defined as follows,

\begin{equation}
\small
\mathcal{L}=\lambda \underbrace{\mathbb{E}_{x \sim p_{x}}\left[-\log _2 p_{\hat{y}}(\hat{y})\right]}_{\text {rate (latents) }}+\underbrace{\mathbb{E}_{x \sim p_{x}}\left[-\log _2 p_{\hat{z}}(\hat{z})\right]}_{\text {rate (hyper-latents) }}+\underbrace{\cdot \mathbb{E}_{x \sim p_{x}}\|x-\hat{x}\|_2^2}_{\text {distortion }}.
\end{equation}

\subsection{Pretrain VAEformer Reconstruction}
\label{sec:pretain}

\begin{figure}
  \centering      \includegraphics[width=\linewidth]{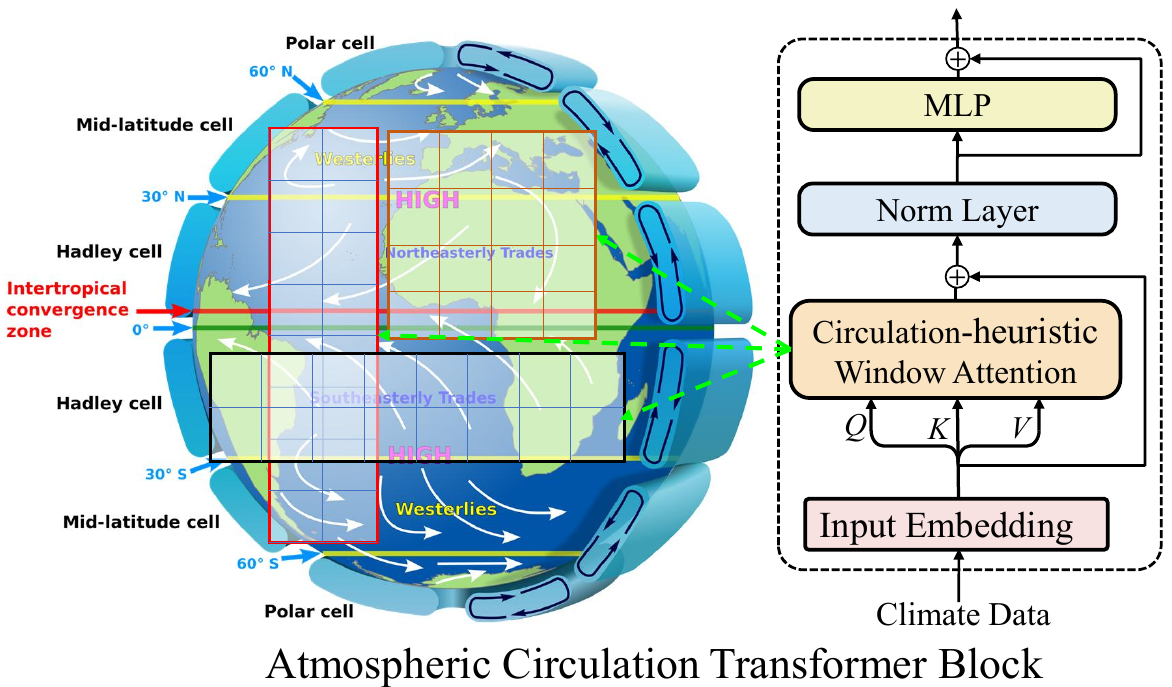}
    \caption{The architecture of the proposed atmospheric circulation transformer block, which performs different shapes of window attention to simulate different kinds of circulation.  
    }
    \label{fig:ACT_block}
    \vspace{-16pt}
\end{figure}

\textbf{Atmospheric Circulation Transformer Block.} 
The transformer network is well-known for capturing global attention among different tokens~\cite{devlin2018bert,radford2018improving,dosovitskiy2020image,peebles2022scalable,radford2021learning}. This characteristic is also suitable for weather data compression as the atmosphere system is chaotic and its long-distance modelling ability can help capture some natural effect, such as butterfly effect. However, one issue is that the complexity increases quadratically with the length of the sequence. Here, we propose the ACT block to decrease complexity while utilizing it for climate data compression. Unlike the existing shift-window transformer~\cite{koyuncu2022contextformer,zhu2021transformer,zou2022devil,liu2023learned}, ACT embeds novel window-attention based on the atmospheric circulation. Fig.~\ref{fig:ACT_block} shows atmospheric circulation is the large-scale movement of air and ocean circulation. Due to the different patterns of atmospheric motion in the east-west and north-south directions, we propose east-west and north-south window attention in addition to the commonly used square-window attention. They can help our compression model better capture atmospheric patterns.
 
ACT performs window-based attention. Thus, we can build an efficient transformer for climate data compression by stacking ACT blocks. As shown in Fig.~\ref{fig:ACT_block}, there are three alternative shapes of window attention in ACT, and we execute them in different blocks. For each window, tokens are flattened into a sequence and perform the multi-head attention mechanism~\cite{vaswani2017attention}, which first generates three learned variables: $Q$ (Query), $K$ (Key), and $V$ (Value) for each token, 
followed by calculating the $K, Q, V$ in the following way,

\begin{equation}
\operatorname{Attn}(Q, K, V)=\operatorname{softmax}\left(\frac{Q K^T}{\sqrt{d_k}}\right) V,
\end{equation}

Defining the $Q$, $K$, and $V$ calculation as a single head, the multi-head attention strategy makes each head $i$ have its projection matrix $W_i^Q$, $W_i^K$, and $W_i^V$. We can obtain multi-head attention as follows,

\begin{equation}
\begin{array}{cc}
     &\operatorname{MultiHead}(Q, K, V)=\operatorname{Concat}\left(\operatorname{H}_1, \ldots, \operatorname{H}_{\mathrm{h}}\right) W^O  \\
     & \operatorname{H}_{i}=\operatorname{Attn}\left(Q W_i^Q, K W_i^K, V W_i^V\right) \\
\end{array},
\end{equation}
where $W_i^Q \in \mathbb{R}^{d_{\text {M}} \times d_k}, W_i^K \in \mathbb{R}^{d_{\text {M }} \times d_k}, W_i^V \in \mathbb{R}^{d_{\text {M}} \times d_v}, W^O \in \mathbb{R}^{h d_v \times d_{\text {M }}}$, $d_M$ is the length of each token, $d_k, d_v$ is the length of each head in Query/Key and Value, generally $d_k=d_v$. $W^{O}$ is the MLP function that maps the attention output with dimension $d_M$.  

\textbf{Transformer-Based Variational Auto-Encoder.} In statistics, variational inference (VI)~\cite{kingma2013auto} is a technique to approximate complex distributions. Most VAE methods~\cite{rombach2022high,ramesh2021zero,kingma2021variational,sohn2015learning} are built upon CNNs. Instead, we here introduce a transformer-based encoder-decoder architecture for variational inference. We aim to approximate $p(y|x)$ by a Gaussian distribution $q_x(y)$, whose mean and covariance are produced by two functions, $\boldsymbol{\mu}_{x}=g(x)$ and $\boldsymbol{\sigma}_{x}=h(x)$ as shown in Fig.~\ref{fig:net}. Thus, our latent encoder models the posterior distribution as $ q_x(y) \equiv \mathcal{N}(g(x), h(x)), g \in G, h \in H$. $H$ and $G$ are a family of candidates for variational inference and the objective is to find the best approximation among this family by optimizing the functions $g$ and $h$  to minimize the Kullback-Leibler (KL) divergence between the approximation and the target $p(y|x)$ (\emph{please see supplementary for the derivation}), 
\begin{equation}
\begin{aligned}
\left(g^*, h^*\right) & =\underset{(g, h) \in G \times H}{\arg \min } K L\left(q_{x}(y), p(y \mid x)\right) \\
& =\underset{(g, h) \in G \times H}{\arg \max }\left(\mathbb{E}_{y \sim q_x}\left(-\frac{\|x-f(y)\|^2}{2 c}\right)-K L\left(q_x(y), p(y)\right)\right),
\end{aligned}
\label{eq:varinfer}
\end{equation}

Eq. \ref{eq:varinfer} assumes the function $f$ is known and fixed. We can approximate the posterior $p(y|x)$ using the variational inference technique. However, in practice, we use the reconstruction transformer decoding process  $\hat{x}=f(y)$ as the mapping function as shown in Fig.~\ref{fig:net}. Our goal is to find a promised encoding-decoding scheme, whose latent space is regular enough to be used for reconstructing the input data. Indeed, as $p(y|x)$ can be approximated from $p(y)$ and $p(x|y)$, and as $p(y)$ is an assumed simple standard Gaussian, the only two levers we have to make optimizations are the parameter $c$ (the variance of the likelihood) and the function $f$ (the mean of the likelihood), Gathering all the pieces together, we are looking for optimal $f^*$, $g^*$ and $h^*$ such that, 

\begin{equation}
\left(f^*, g^*, h^*\right)=\arg \max \left(\mathbb{E}_{y \sim q_x}\left(-\frac{\|x-f(y)\|^2}{2 c}\right)-K L\left(q_x(y), p(y)\right)\right),
\label{eq:objective}
\end{equation}
This objective intuitively shows the optimization elements of VAEs: 
the reconstruction error between $x$ and $\hat{x}$
and the regularisation term given by the KL divergence between $q_x(y)$ and $p(y)$. Constant $c$ rules the balance between the two previous terms. Note that $y$ is sampled from the distribution predicted by the encoder. The sampling process has to be expressed in a way that allows the error to be back-propagated through the network. We use the reparametrization strategy as in \cite{kingma2013auto} to make the gradient descent possible during the training. In fact, our produced $y$ will naturally become a random variable following a Gaussian distribution with mean $\boldsymbol{\mu}_x$ and with covariance $\boldsymbol{\sigma}_x$ then it can be expressed as,
\begin{equation}
y= \boldsymbol{\mu}_x  + \boldsymbol{\sigma}_x \odot \boldsymbol{\epsilon},   \quad \boldsymbol{\epsilon} \sim \mathcal{\mathcal{N}}(\mathbf{0}, I),
\end{equation}

Finally, the theoretical expectancy in Eq.~\ref{eq:objective} is replaced by a more or less accurate Monte-Carlo approximation~\cite{duane1987hybrid}. The pretraining objective in this section consists of a reconstruction term and a regularisation term, 
\begin{equation}
\mathcal{L}= \frac{1}{2}\|\boldsymbol{x}-\hat{\boldsymbol{x}}\|^2 + \frac{1}{2} \sum_{i=1}^d\left(-\log \sigma_i^2+\mu_i^2+\sigma_i^2-1\right).
\end{equation}

\subsection{Fine-tune Entropy Model}
Sec.~\ref{sec:pretain} guides us to pre-train a conditional generative model, which gives a well-predictable latent representation $y$. Unless specified, we load the pre-trained parameters to optimize the learning of the mean scale hyperprior introduced in Sec.~\ref{sec:compreesor} to estimate the distribution of quantized latent representation $\hat{y}$ for a higher compression ratio. During this phase, we froze the latent transformer encoder and only fine-tune the parameters of the other learnable components, which makes full use of the pre-trained model, resulting in the best compression performance and a relatively lower training cost. 

\textbf{Optimization of our Entropy Model.} 
During the optimization procedure, we follow the strategy as in~\cite{minnen2018joint}. Here, the entropy model needs to approximate the quantization operation by an additive uniform noise because the round operation in quantization is not derivable, which ensures a good match between encoder and decoder distributions of both the quantized latent, and continuous-valued latent subjected to additive uniform noise. The variational inference of $\hat{y}$ is a Gaussian distribution $\mathcal{N}(\boldsymbol{\mu}, \boldsymbol{\sigma})$ convolved with a unit uniform distribution, 
\begin{equation}
p_{\hat{y}}(\hat{y} \mid \hat{z}, \boldsymbol{\theta}_{\boldsymbol{h}_s}, \boldsymbol{\theta_{ep}})=\prod_{i=1}\left(\mathcal{N}\left(\mu_i, \sigma_i^2\right) * \mathcal{U}(-\frac{1}{2},\frac{1}{2})\right)\left(\hat{y}_i\right).
\end{equation}

Regarding the effectiveness of using such a two-phase optimization strategy, please see the ablation study in Sec.~\ref{sec:ablation} for a detailed comparison. 
\emph{Some model details are provided in the supplementary.}

\begin{table*}[t]
	\centering
	\caption{Ablation results of Weighted RMSE and the overall MSE between our compressed data and the original data. $\lambda$ is set to $1e-2$ for all experiments. ``Train from Scratch'' represents models that are trained from the initial state. ``CNN'' means the entropy model is a convolutional neural network.}
	\label{tab:ablation}
	\begin{tabular}{@{}lccccc|cccccc|cc@{}}
		\toprule
		&\multirow{2}{*}{Method} &\multirow{2}{*}{\makecell[c]{Sampled \\ $y$}} & \multirow{2}{*}{\makecell[c]{Pretrain \\ model}} & \multirow{2}{*}{\makecell[c]{Entropy\\ Model}} & \multirow{2}{*}{\makecell[c]{Frozen \\ Encoder}}& \multicolumn{6}{c|}{Weighted RMSE $\downarrow$} &\multirow{2}{*}{\makecell{Overall \\ MSE  $\times 100$}}& \multirow{2}{*}{\makecell[c]{Comp.\\ Ratio}} \\
		\cmidrule{7-12} &&&&& & z500 & t850& v10 & u10 & t2m & msl & & \\
		\midrule
		  1)& Pretrained VAEformer  & \checkmark & \checkmark & \xmark  & \xmark & 26.15 &0.58 & 0.45 &0.49 &0.74& 28.02&0.81&16.0 \\
            \midrule
		2)& Train from Scratch   & \xmark     & \xmark     & \checkmark  & \xmark & 77.38 &1.21 & 1.10 &1.15&1.62&88.95&3.10 &\textbf{1845.0}\\
            3)&Train from Scratch   & \checkmark &\xmark      &CNN   & \xmark & 96.10 & 1.29& 1.10&1.16& 1.96&103.41&3.13&713.6  \\
            4)&Train from Scratch   & \checkmark &\xmark      &\checkmark   & \xmark  & 34.84 & 0.731&0.62&0.63&0.94&37.80 &1.22&201.9
            \\
            5)&pretrain$+$AE$+$AD  & \checkmark &\checkmark      &\xmark   & \xmark & 42.14 & 0.75&0.68&0.71&0.98&48.87&1.47&22.6 \\
		  6)&Fine-tuning  & \checkmark &\checkmark      &\checkmark   & \xmark & \textbf{31.23} & \textbf{0.65}  & \textbf{0.59} &0.62&0.80&36.73 &1.15&220.8\\
		7)&Fine-tuning  & \checkmark &\checkmark  &\checkmark & \checkmark &31.89 & 0.66  &0.80& \textbf{0.59} &\textbf{0.62} &\textbf{36.69} &\textbf{1.15} &221.1\\
	\bottomrule
	\end{tabular}
\end{table*}
\section{Experimental Results}
\subsection{Dataset}
\label{era5:128}
To economically evaluate the compression performance of all models, we sample a low-resolution ($128\times256$) version of the most popular ERA5~\citep{hersbach2020era5} dataset, which is a global atmospheric reanalysis dataset produced by the European Centre for Medium-Range Weather Forecasts. It provides comprehensive information about the earth's climate data. In this paper, we select 5 atmospheric variables (each with 13 pressure levels) and four surface variables, a total of 69 dimensions in the channel. Specifically, the atmospheric variables are geopotential ($z$), relative humidity ($r$), zonal component of wind ($u$), meridional component of wind ($v$), and air temperature ($t$), whose 13 sub-variables at the different vertical level are presented by abbreviating their short name and pressure levels (\emph{e.g.}, z500 denotes the geopotential height at a pressure level of 500 hPa). The four surface variables are 2-meter temperature (t2m), 10-meter u wind component (u10), 10-meter v wind component (v10), and mean sea level pressure (msl). 

The dataset is split into training/validation/testing sets according to years. i.e., the training set comprised of 1979-2015 years, the validation set that includes 2016-2017, and the testing set with 2018-year data. The sizes are 455.0 GB, 24.6 GB, and 12.3 GB, respectively. To the best of our knowledge, we are the first work to compress such a large-scale climate dataset with learning methods.

\subsection{Implementation Details}
\label{cfg:vaeformer}
The pretraining and fine-tuning require 60 epochs and 30 epochs, respectively, utilizing 8 Tesla A100 GPUs. The batch size is 32 and the optimization is achieved through the AdamW~\cite{loshchilov2017decoupled} algorithm. The initial learning rates are set at $2e-4$ and $5e-5$ for pretraining and fine-tuning, respectively. We use a linear learning rate decay schedule to warm up the learning rate from 0 to the initial value, followed by a cosine decay learning rate schedule. The best-performance model is selected based on its performance on the validation set. 

\vspace{-3mm}
\subsection{Ablation Study}
\label{sec:ablation}
To pick up the performant configurations, we perform a comprehensive ablation study by selectively disabling some modules of the model or adjusting the training strategy. Comparing the resulting weighted RMSE loss between compressed data and original data, we find that the pre-trained model has the best performance on restoring the weather data. However, the compression ratio is relatively lower because it is only with the downsampling rate in the latent encoder as its compression ratio. From Table~\ref{tab:ablation}, our primary findings are:

\textbf{Transformer is better for weather data.} By comparing experiments 2) and 3), the hyperprior transformer encoder is replaced by CNN in 3). Under the same hyper-parameters, CNN obtains a higher compression ratio of 713.6$\times$, but the reconstruction results are significantly poor. As weather data compression prioritizes accuracy, we recommend using the transformer-based network.

\textbf{Pretrain is helpful.} With different configurations, the ``train from scratch'' models are consistently underperforming than the pretrain+fine-tuning models, which verify the pretraining strategy we proposed to decouple the task complexity is an effective means.  

\textbf{Entropy model works for weather data.} Experiment 5) in Table~\ref{tab:ablation} shows the compression ratio can be slightly improved (from 16.0$\times$  to 22.6$\times$) when directly applying the arithmetic coding and decoding to the pre-trained model. However, the reconstruction results drop substantially compared with experiment 1). While fine-tuning the pre-trained models, the compression ratio further increases to about 220$\times$, and the reconstruction errors decrease to acceptable ranges.

\textbf{Frozen encoder is a better fine-tuning method.} 
Experiments 6) and 7) compare the difference of freezing the encoder in the pre-trained model, showing frozen operation neither affects the reconstruction nor increases the compression ratio. So, we finally fixed it for memory-saving and computation-efficient fine-tuning.

\subsection{Comparision with Other NIC Methods}

\begin{table*}[t]
	\centering
	\caption{The compression performance between the proposed VAEformer and the NIC methods. bpsp is the bits per pixel from~\cite{mentzer2019practical}. The encoding and decoding times are calculated in Linux system with Intel(R) Xeon(R) Gold 6248R CPU @ 3.00GHz and NVIDIA GPU A100-SXM4-80GB.  * denotes that the original model is scaled to have comparable parameters as VAEformer.}
	\label{tab:comparison_sota}

	\begin{tabular}{@{}lclclccccccccccc@{}}
		\toprule
		\multirow{2}{*}{Method} &\multirow{2}{*}{$\lambda$} & \multicolumn{6}{c}{Weighted RMSE $\downarrow$} &\multirow{2}{*}{\makecell[c]{Overall\\ MSE $\times$ 100}}&\multirow{2}{*}{\makecell[c]{Comp.\\ Ratio}} &\multirow{2}{*}{\makecell[c]{bpsp}} & \multirow{2}{*}{\makecell[c]{encoding\\time (s)}}& \multirow{2}{*}{\makecell[c]{decoding\\time (s)}}\\
		\cmidrule{3-8} & & z500 & t850& v10 & u10 & t2m & msl  \\
            \midrule
            Mean (reference)&-&  54117& 275&0.19&-0.05&279&100957&$\downarrow$&$\uparrow$&$\downarrow$&$\downarrow$&$\downarrow$\\
            \midrule
	cheng2020~\cite{cheng2020learned}  &1.25e-8&121.61&1.53&1.31&1.36&2.17&110.72&4.06&229.8 
 &0.139&0.299&0.603\\	  
    
    bmsj2018~\cite{balle2018variational}  & 5e-8 & 69.95
 &1.26  &1.02&1.07&1.7&81.53&2.58&242.5&0.132 & \textbf{0.011} &\textbf{0.015}\\
   
    mbt2018~\cite{minnen2018joint}    & 1e-7   & 68.75  &1.26 & 1.02&1.07&1.68&81.45&2.59&265.3 &0.126 &0.306 &0.640 \\
    Inv2021~\cite{xie2021enhanced}    & 1e-3  & 101.01	&1.09	&0.80	&0.85	&1.37	&72.83	&1.72	&202.5	&0.158 &1.189 &	2.536 \\ 
    ELIC2022-467M$^*$\cite{he2022elic}	&1e-6	&47.74	&0.99	&0.83	&0.86	&1.17	&56.29	&1.89	&\textbf{285.6}	&\textbf{0.112} &0.074	&0.095 \\
    TCM2023-300M$^*$~\cite{liu2023learned}	&1e-5	&49.87	&0.88	&0.77	&0.80	&1.06	&48.90	&1.65	&256.0	&0.120 &0.095	&0.109 \\
            VAEformer   & 10&\textbf{33.13} &\textbf{0.66}  &\textbf{0.59}&\textbf{0.62}&\textbf{0.81}&\textbf{37.42} & \textbf{1.15} &229.6 &0.139 &0.098 &0.035 \\
 \bottomrule
\end{tabular}
 \vspace{-2mm}
\end{table*}

\begin{figure}
  \centering     
  \includegraphics[width=\linewidth]{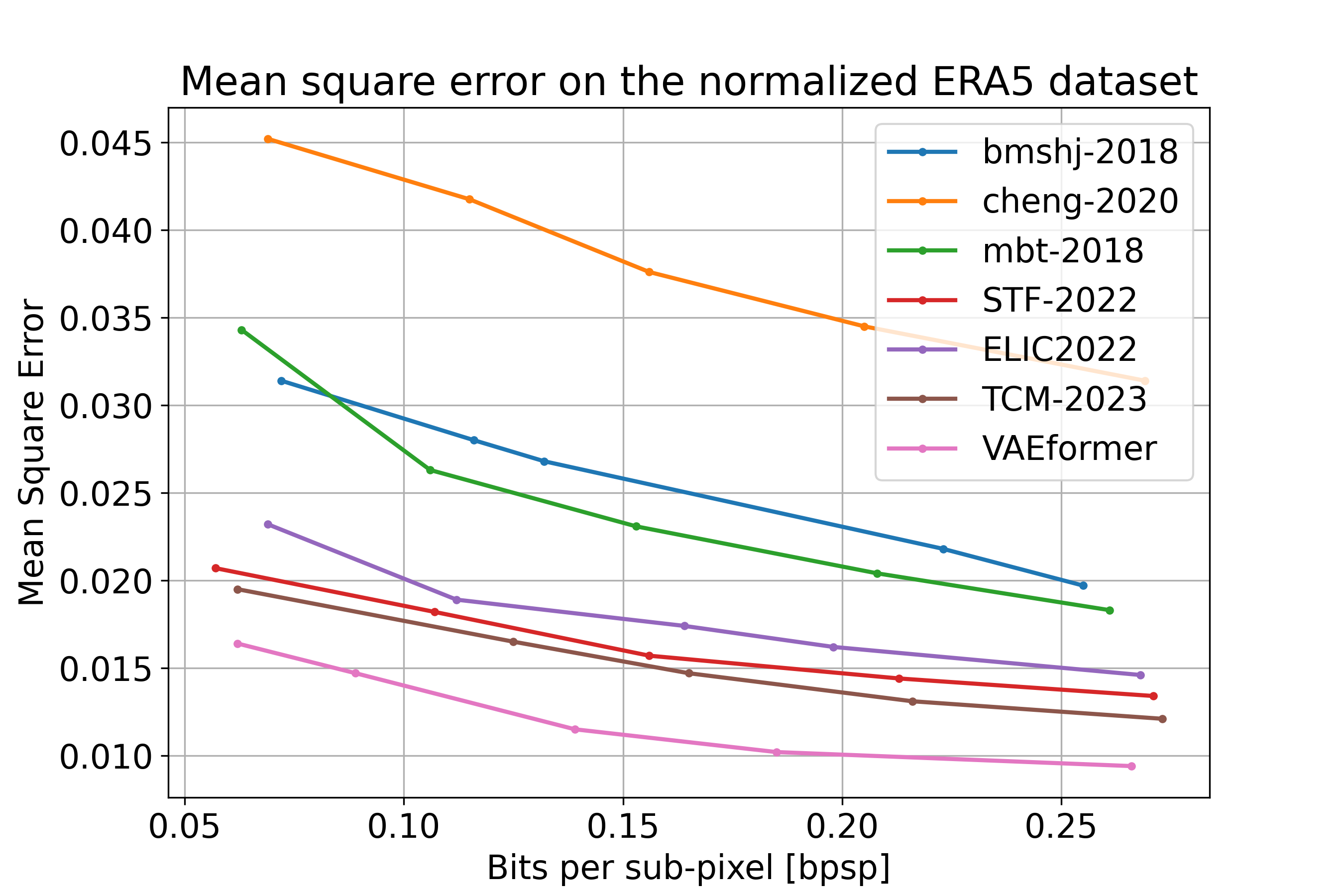}
    \caption{Rate-distortion (\emph{i.e.}, bit-rates vs MSE) performance comparison between VAEformer and previous NIC methods on ERA5. Here distortion degree is based on MSE.}
    \label{fig:RD_Curve}
\end{figure}


Table~\ref{tab:comparison_sota} presents a comparison of our compression performance with state-of-the-art NIC methods for each year: bmsj2018~\cite{balle2018variational}, mbt2018~\cite{cheng2020learned}, cheng2020~\cite{cheng2020learned}, Inv2021~\cite{xie2021enhanced}, ELIC2022-467M~\cite{he2022elic}, and TCM2023-300M~\cite{liu2023learned}. We use the code in CompressAI~\footnote{\url{https://github.com/InterDigitalInc/CompressAI}} to re-implement them. For fair comparison, we adjust their $\lambda$ to reach the comparable compression ratios with our VAEformer. To measure the compression results comprehensively, we compute weighted RMSE~\cite{lam2022graphcast} for multi-variables, overall MSE, compression ratio, bits per sub-pixel~\cite{mentzer2019practical}, encoding, and decoding time for each method. It tabulates that the proposed VAEformer comprehensively improves the evaluation targets (z500, t850, u10, v10 \emph{et al.}). The MSE of decompressed data is significantly lower than that of NIC-based methods with a similar compression ratio. VAEformer also shows a faster encoding and decoding time than most NLC methods, which guarantees it is practical for real-world applications.

We also evaluate the effects of our transformer-baed entropy model by calculating the rate-distortion (RD) performance. Fig.~\ref{fig:RD_Curve} shows the RD curves over the ERA5 by using overall MSE as the quality metric, which intuitively depicts that our VAEformer achieves the lowest MSE at different bit-rates compared with the state-of-the-art NIC methods.

\section{Showcase: CRA5 Dataset} 
To facilitate accessible and portable weather research, we compressed hundreds of TBs of ERA5 weather data into a more manageable size less than 1 TB, termed CRA5, using our VAEformer

\textbf{Dataset}: 
We adopt ERA5 dataset with full-resolution data, which has dimensions of $159 \times 721 \times 1440$ and includes 6 atmospheric variables across 25 pressure levels, along with 9 surface variables. Each data instance, stored in float32 precision (32 bits), amounts to approximately 629.73 MB. We retrained our VAEformer model on the dataset spanning 43 years (1979-2021), totalling 226.2 TB. The resulting compressed dataset, named CRA5, occupies only 0.7 TB.

However, as previously mentioned, climate data is continually collected and released every day. To evaluate the sustainability of using our VAEformer for compressing newly incoming data, we adopted unseen data from 2022 and 2023 as our validation and testing sets to assess performance.

\textbf{Training details}:
Here, we set $\lambda$ to $1e-3$ to facilitate an extreme compression training procedure. Most configurations remain unchanged from Section~\ref{cfg:vaeformer}, except for specific architectural hyperparameters. For patch embedding in the convolution layers, we modified the kernel size to $11\times10$, stride to  $10\times10$, and increased the latent dimension to 256. Regarding optimization, we again adopted our two-stage process: reconstruction pretraining for 150K iterations, followed by entropy model fine-tuning for 30K iterations, with a batch size of 8. The training was conducted using 8 Tesla A100 GPUs, spanning four and two days for the above two stages.

\textbf{Evaluation} 
We adopt the test data (2023) to perform the evaluation regarding \emph{Compression Ration} and \emph{Extreme Verification}.

\emph{Compression Ration.} Each evaluation instance, originally sized at 629.73 MB, is compressed into a 2 MB string, achieving a compression ratio of over 300x. This performance is consistent with that observed during the training phase.

\begin{figure*}
  \centering      
  \includegraphics[width=0.92\textwidth]{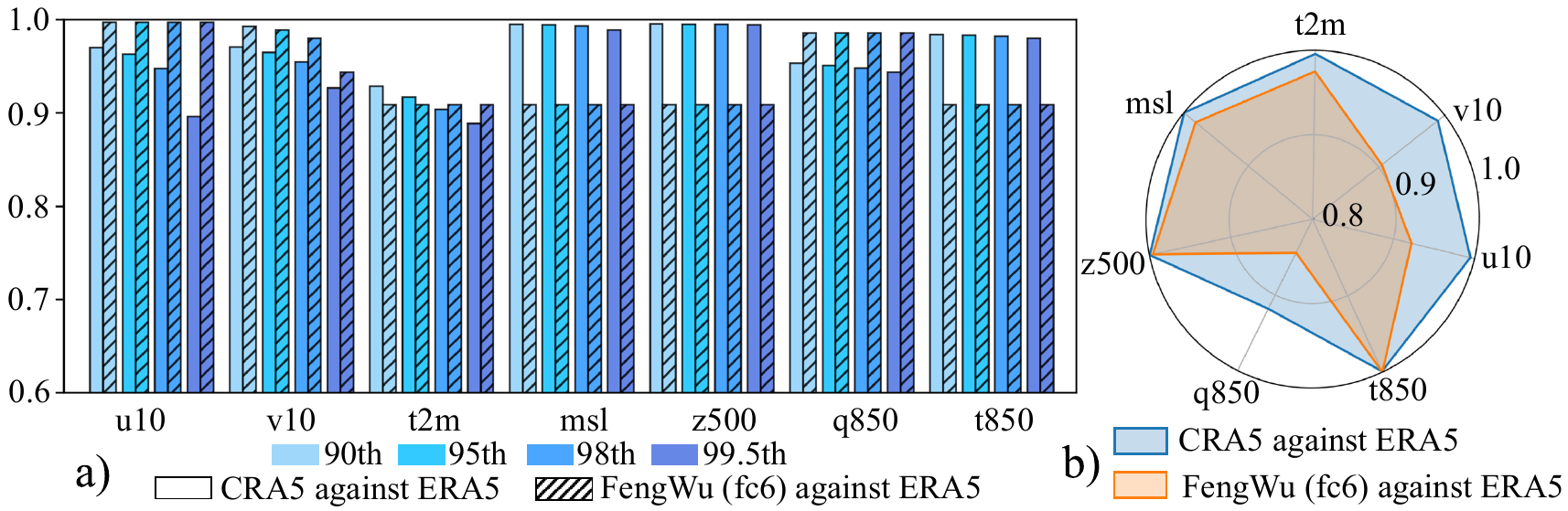}
    \caption{ a) SEDI, the closer to 1 the better. b) RQE$\in [-1,1]$, $< 0$ means underestimating extreme values, $> 0$ means overestimating extreme values. Here we use $1+10*RQE$ to draw the radar figure for intuitive comparison. Thus, the closer to 1 the better. All results are tested on the data in 2023, and ERA5 is used as the target. 
}
    \label{fig:SEDI_RQE}
\end{figure*}

\emph{Extreme Verification.} 
For climate and weather research, extreme events, such as tropical cyclone~\cite{wang2024global} and heat wave, highly depend on extreme values. To check whether CRA5 keeps enough extreme information for climate research, following some extreme weather research~\cite{xu2024extremecast,kurth2023fourcastnet}, we use Symmetric Extremal Dependency Index (SEDI) and Relative Quantile Error (RQE) to measure the extremeness against the raw ERA5. SEDI classifies each pixel into extreme or normal weather with large quantiles ($90\%, 95\%, 98\%$, and $99.5\%$) as thresholds and then calculates
the hit rate. As some key variables shown in Fig.~\ref{fig:SEDI_RQE}, CRA5 achieves a hit rate of more than $90\%\sim100\%$ across all extreme thresholds. Here, we also calculate the SEDI on the prediction results (6-hour lead time) of FengWu~\cite{chen2023fengwu}. Results show that CRA5 yields the comparable SEDI, which ensures the feasibility of training a weather forecasting model with performance similar to FengWu using CRA5. RQE in Fig.~\ref{fig:SEDI_RQE}  also demonstrates CRA5 well recovery extreme values in ERA5.  

\section{Downstream: Weather Forecasting}

\begin{figure*}
  \centering      \includegraphics[width=0.95\textwidth]{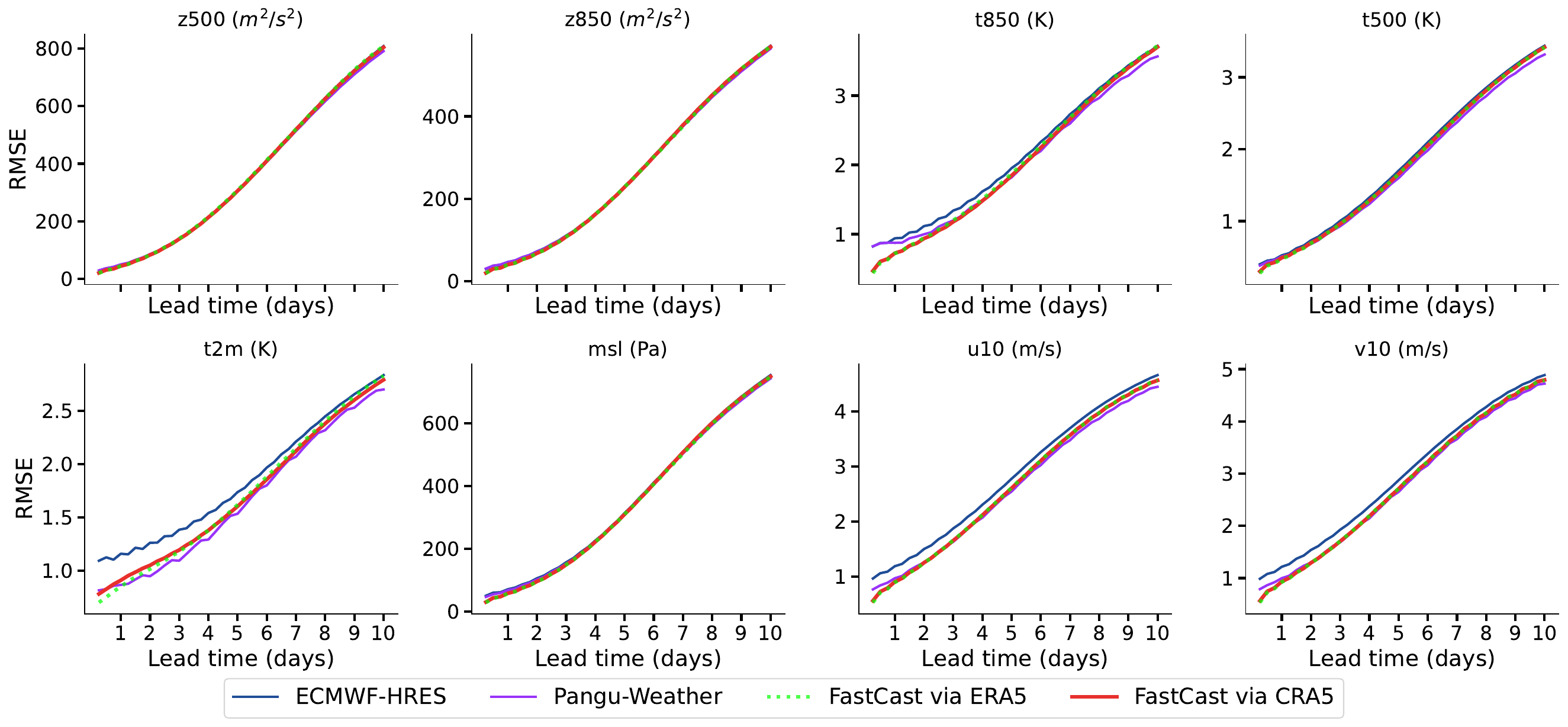}
    \caption{\textbf{The weather forecast skill in 2022 (lower is better).} These subfigures show the latitude-weighted RMSE on different variables when weather forecast models are trained by the CRA5 dataset (\textcolor{red}{red} dashed lines) and ERA5 dataset (\textcolor{green}{green} dotted lines), respectively. The x-axis in each sub-figure represents lead time, at 6-hour steps over 10 days. The results show the forecasting model driven by the CRA5 can achieve comparable forecast skills compared with the model driven by the ERA5.}
    \label{fig:weather_forecast}
\end{figure*}

While pursuing a higher compression ratio, we also need to maintain the utility of the compressed data. Thus, we conduct downstream experiments on the compressed data as another metric to evaluate our compression performance. For atmospheric research, some data-driven weather forecasting models~\cite{chen2023fengwu, dryden2021clairvoyant} have suffered the I/O bound during training and transmitting. Alternatively, compressed climate data would be a solution to tackle this issue~\cite{huang2022compressing}.

In this section, we develop a global numerical weather prediction model and evaluate the forecast skill by respectively learning from CRA5 and ERA5. 
This experiment almost takes the same configuration as the current state-of-the-art NWP model~\cite{han2024fengwu}. 
Moreover, to support weather research groups with limited resources, we also developed a lightweight forecasting model that incorporates the proposed ACT block, termed \emph{FastCast}. 
For training, we select the important 69 variables as described in Sec.~\ref{era5:128} from both original and compressed data with the size of $69\times721\times1440$. Fig.~\ref{fig:weather_forecast} illustrates the predictive performance comparison of FastCast when training on compressed CRA5 and original ERA5. The comparison is carried out over 8 variables at 6-hour intervals, 10-days lead time. 

The results clearly show that CRA5 does not lose the ability to train an advanced forecasting model. It achieved performance almost on par with the model trained on the original ERA5. Even with the lead time increasing, they still have the same forecast skills. Apart from that, we also compare FastCast with the most advanced physical-based model, ECMWF's High RESolution forecast (ECMWF-HRES~\cite{81380}), and the most influential AI-based weather model, Pangu-Weather~\cite{pathak2022fourcastnet}. The curves show that it achieved a performance comparable to that of Pangu-Weather and ECMWF-HRES, even surpassing them in certain variables and lead times. This evidence suggests that the CRA5 dataset is still suitable for scientific research. However, its optimal size will facilitate more researchers to engage in climate research.

\section{Conclusion}
In this work, we exploit advanced NIC techniques to make two significant contributions to the weather research community. 
First, we propose a transformer-based practical weather data compression framework VAEformer, which adopts an efficient variational transformer to generate the latent feature via variational inference and hyper-prior information for performing entropy coding. Leveraging the powerful ACT Transformers and our two-phase optimization strategy, our framework is capable of compressing the climate into a small compact representation and significantly decreasing the reconstruction error compared with other conventional NIC methods while having a comparable compression ratio. Second, 
we have created a compressed weather dataset, CRA5, which requires only 1/300 the storage space of its original version, ERA5, while retaining most of the crucial information. This ensures its usability for downstream analytical tasks, such as weather forecasting. Looking ahead, we aim to expand our method to broader applications. For instance, we plan to explore how to effectively apply the VAEformer for lossless compression of climate data.









\bibliographystyle{ACM-Reference-Format}
\bibliography{sample-base}









\end{document}


\settopmatter{printacmref=false} 
\renewcommand\footnotetextcopyrightpermission[1]{} 
\pagestyle{plain} 





\newcommand{\MyMapTemplatePrefixtb}[5]{\expandafter#1\csname#4#5\endcsname{#2{#3{#5}}}} 
\forcsvlist{\MyMapTemplatePrefixtb {\def} {\tilde}{\mathbf}{t}} {A,B,C,D,E,F,G,H,I,J,K,L,M,N,O,P,Q,R,S,T,U,V,W,X,Y,Z}  
\forcsvlist{\MyMapTemplatePrefixtb {\def} {\tilde}{\mathbf}{t}} {0,1,a,b,c,d,e,f,g,h,i,j,k,l,m,n,o,p,q,r,s,u,v,w,x,y,z}  

\newcommand{\MyMapTemplateNoPrefix}[3]{\expandafter#1\csname#3\endcsname{#2{#3}}}
\forcsvlist{\MyMapTemplateNoPrefix {\def} {\mathbf} } {0,1,a,b,c,d,e, f, g, h, i, j, k, l, m, n, o, p, q, r, u, v, w, x, y, z} 
\forcsvlist{\MyMapTemplateNoPrefix {\def} {\mathbf} } {A,B,C,D,E,F,G,H,I,J,K,L,M,N,O,P,Q,R,S,T,U,V,W,X,Y,Z}  

\def\ba{\bm{\alpha}}
\def\bb{\bm{\beta}}
\def\bd{\bm{\delta}}
\def\bt{\bm{\theta}}
\def\bz{\bm{\zeta}}

\def\bP{\bm{\Phi}}
\def\bQ{\bm{\Psi}}
\def\bS{\bm{\Sigma}}

\def\bbR{{\mathbb R}}
\def\bbE{{\mathbb E}}

\def\tr{\mbox{tr}}
\def\Pr{{P}}

\def\etal{\emph{et al.}\@\xspace}
\def\ie{\emph{i.e.}\@\xspace}
\def\eg{\emph{e.g.}\@\xspace}
\def\resp{\emph{resp.}\@\xspace}

\definecolor{rowblue}{RGB}{220,230,240}

\definecolor{linkcol}{RGB}{233, 4, 141}
\newcommand{\linkcol}[1]{\textcolor{linkcol}{#1}}
\definecolor{xycolor}{RGB}{60, 120, 216}
\definecolor{xycolor}{HTML}{0071bc}
\newcommand{\xycolor}[1]{\textcolor{xycolor}{#1}}
\definecolor{wcolor}{RGB}{103, 78, 167}
\newcommand{\wcolor}[1]{\textcolor{wcolor}{#1}}
\definecolor{dcolor}{RGB}{166, 77,21}
\newcommand{\dcolor}[1]{\textcolor{dcolor}{#1}}
\definecolor{gcolor}{RGB}{204, 102, 153}
\newcommand{\gcolor}[1]{\textcolor{gcolor}{#1}}
\definecolor{tcolor}{RGB}{34,139,34}
\newcommand{\tcolor}[1]{\textcolor{tcolor}{#1}}
\definecolor{iterc}{RGB}{91,196,159}
\newcommand{\iterc}[1]{\textcolor{iterc}{#1}}
\definecolor{epochc}{RGB}{96,172,252}
\newcommand{\epochc}[1]{\textcolor{epochc}{#1}}
\definecolor{eicolor}{RGB}{153, 51, 102}
\newcommand{\eicolor}[1]{\textcolor{eicolor}{#1}}
\newcommand{\maskcolor}[1]{\textcolor{orange}{#1}}
\newcommand{\outsizesRaw}[4]{\multirow{#4}{*}{\(\begin{array}{c}  \text{#1$\times$#2\x #3}\\[-.1em]  \end{array}\)}}
\newcommand{\outsizesRawD}[5]{\multirow{#5}{*}{\(\begin{array}{c}  \text{#1\x#2\x#3\x#4}\\[-.1em]  \end{array}\)}}

\newcommand{\blockatt}[4]{\multirow{3}{*}{\(\left[\begin{array}{c}\text{CHWA(\wcolor{#1})}\\[-.1em] \text{CHWA(\wcolor{#2})}\\[-.1em]\text{CHWA(\wcolor{#3})}\end{array}\right]\)$\rightarrow$#4}
}
\title[CRA5:Extreme Compression of ERA5 for Portable Global Climate and Weather Research ]{CRA5: Extreme Compression of ERA5 for Portable Global Climate and Weather Research via an Efficient Variational Transformer (Supplementary Materials)}

\maketitle

\section{Index}
This supplementary file provides additional details of our VAEformer from the following aspects:
\begin{enumerate}

  \item Derivations of variational inference is in $\S$  \ref{sec:A}.
  \item Detailed network architectures are in $\S$ \ref{sec:B}.
  \item More experimental results and analysis are in $\S$ \ref{sec:extra_exp}.
  \item Visualization of the reconstructed samples is in $\S$~\ref{sec:vis}.
  \item Discussion of Limitations is in $\S$~\ref{sec:limit}.
  \item Discussion of Potential negative societal impact is in  $\S$~\ref{sec:potential}.

\end{enumerate}

\section{Variational Inference Formulation}
\label{sec:A}
\textbf{Optimization Objective}. Eq. 6 in our main manuscript defines the Kullback-Leibler divergence $KL(\cdot)$ between the approximation $q_x(y)$ and the target $p(y|x)$. The optimization of variational inference aims to produce the optimal mapping functions $g^*$ and $h^*$ (please see Section~3.1 for more details) such that, 

\begin{equation}
\begin{aligned}
\left(g^*, h^*\right) & =\underset{(g, h) \in G \times H}{\arg \min } K L\left(q_x(y), p(y \mid x)\right) \\
& =\underset{(g, h) \in G \times H}{\arg \min }\left(\mathbb{E}_{y \sim q_x}\log q_x(y)-\mathbb{E}_{y \sim q_x} \log \frac{p(x \mid y) p(y)}{p(x)}\right) \\
& = \underset{(g, h) \in G \times H}{\arg \min} \left( \mathbb{E}_{y \sim q_x}\log q_x(y) -\mathbb{E}_{y \sim q_x} \log p(y)- \right.\\ & \hspace{2cm} \left. \mathbb{E}_{y \sim q_x}\log p(x \mid y)+\mathbb{E}_{y \sim q_x}\log p(x)\right) \\
& =\underset{(g, h) \in G \times H}{\arg \max }\left(\mathbb{E}_{y \sim q_x}(\log p(x \mid y))-K L\left(q_x(y), p(y)\right)\right) \\
& =\underset{(g, h) \in G \times H}{\arg \max }\left(\mathbb{E}_{y \sim q_x}\left(-\frac{\|x-f(y)\|^2}{2 c}\right)-K L\left(q_x(y), p(y)\right)\right).
\end{aligned}
\label{eq:varinfer}
\end{equation}

To this end, we want to maximize the last equation as in Eq.~6 in our main manuscript. Specifically, its first term, $\mathbb{E}_{y \sim q_x}\left(-\frac{\|x-f(y)\|^2}{2 c}\right)$, is to maximize the logarithmic likelihood function, while its second term, $K L\left(q_x(y), p(y)\right)$, aims to minimize the distance between $q_x(y)$ and $p(y)$. The term, $K L\left(q_x(y), p(y)\right)$, on the right means there is a $x$ that can sample a $y$ based on the distribution $q_{x}$. This procedure maps $x$ into $y$ with a parameterized encoder. 

In the second to last equation, $p(x|y)$ represents that we generate $\hat{x}$ on the basis of $y$, namely, ``reconstruction" in our compression method. Specifically, its first term represents the expectation of the likelihood function to measure the ``reconstruction error" when decoding the latent feature $y$ to a $\hat{x}$ with a given $x$. If the expectation is higher, we are more likely to produce a better latent representation $y$ for reconstructing a better $\hat{x}$. Its second term $K L\left(q_x(y), p(y)\right)$ is a regularization term between the posterior distribution $q_{x}(y)$ of $y$ and the prior distribution $p(y)$.

\textbf{Implementation of $K L\left(q_x(y), p(y)\right)$}. 
As illustrated in Figure~1 of our main manuscript, we use the transformer encoder to estimate the mean $\mu_{x}$ and variance $\sigma_{x}$ of the input data, which means each sample $x$ corresponds to its mean vector and variance vector and intuitively forms a normal distribution for this sample. Such distribution allows adjustment of the variance $\sigma_{x}$ slightly, so that even if we randomly sample from such distribution, we can still obtain sample $\hat{x}$ close to the real sample $x$. 

In the extreme case that there is no KL divergence term, the predicted variance would degenerate to 0 to generate samples similar to the real sample. The process therefore becomes an ordinary auto-encoder and lose the ability to generate new samples as in most existing NIC methods. Hence, to avoid such a case, we then have to maintain a small variance. Consequently, we adopt the KL divergence. Specifically, we can make $p(y|x)$ approaching $\mathcal{N}(0,1)$, which constrains the variance not equal to 0. Therefore, let $p_{y}\sim \mathcal{N}(0,1)$,  we will have $KL\left(q_x(y), p(y)\right)=KL(\left(q_x(y),\mathcal{N}(0,1)\right))$, 

\begin{equation}
\begin{aligned}
& K L\left(q_x(y) \| \mathcal{N}(0, I)\right) \\
= & \int \frac{1}{\sqrt{2 \pi \sigma^2}} e^{-\frac{(x-\mu)^2}{2 \sigma^2}} \ln \frac{\frac{1}{\sqrt{2 \pi \sigma^2}} e^{-\frac{(x-\mu)^2}{2 \sigma^2}}}{\frac{1}{\sqrt{2 \pi}} e^{-\frac{x^2}{2}}} \mathrm{~d} x \\
= & \int \frac{1}{\sqrt{2 \pi \sigma^2}} e^{-\frac{(x-\mu)^2}{2 \sigma^2}} \ln \frac{1}{\sqrt{\sigma^2}} \times e^{\frac{x^2}{2}-\frac{(x-\mu)^2}{2 \sigma^2}} \mathrm{~d} x \\
= & \int \frac{1}{\sqrt{2 \pi \sigma^2}} e^{-\frac{(x-\mu)^2}{2 \sigma^2}}\left[-\frac{1}{2} \ln \sigma^2+\frac{1}{2} x^2-\frac{1}{2} \frac{(x-\mu)^2}{\sigma^2}\right] \mathrm{d} x \\
= & \frac{1}{2} \int \frac{1}{\sqrt{2 \pi \sigma^2}} e^{-\frac{(x-\mu)^2}{2 \sigma^2}}\left[-\ln \sigma^2+x^2-\frac{(x-\mu)^2}{\sigma^2}\right] \mathrm{d} x \\
= & \frac{1}{2}\left(-\ln \sigma^2+\mathbb{E}\left[x^2\right]-\frac{1}{\sigma^2} \mathbb{E}\left[(x-u)^2\right]\right) \\
= & \frac{1}{2}\left(-\ln \sigma^2+\sigma^2+\mu^2-1\right)
\end{aligned}
\end{equation}

The KL divergence representation introduced above is for a 1-dimensional hidden variable $\y$. Since the normal distribution is isotropic, it is further represented as follows, when the dimension is $d$,
\begin{equation}
\mathcal{L}_{KL} = \frac{1}{2} \sum_{i=1}^d\left(-\log \sigma_i^2+\mu_i^2+\sigma_i^2-1\right).
\end{equation}

\section{Detailed Network Architectures }
\label{sec:B}
As mentioned in \emph{Sec.}~3.2 in the main manuscript, we employ the proposed Atmospheric Circulation Transformer (ACT) blocks as the basic Transformer module in our Encoder-Decoder networks for producing both latent feature and hyper-prior in our VAEformer. 
Here, we specify the detailed architectural configuration for our encoder and decoder as shown in Table~\ref{tab:arch1} and Table~\ref{tab:arch2}, respectively. When stacking the transformer blocks, we recursively use the Circulation-Heuristic Window Attention (CHWA) for efficient computation.

\def\x{$\times$}
\begin{table}[h]
    \scriptsize
    \centering
    \tablestyle{1pt}{1.08}
    \begin{tabular}{c|c|c}
    \textbf{Stage} & \textbf{\makecell[c]{Latent-Reconstruction\\ Encoder-Decoder}} & \textbf{Output Sizes}  \\  
    \hline
    input data & B\x\wcolor{69}\x\xycolor{128}\x\xycolor{256}  &   -  \\
    \hline
    patch embed & Conv \wcolor{1024}\x\wcolor{V}\x\xycolor{4}\x\xycolor{4} & B\x\wcolor{1024}\x\xycolor{32}\x\xycolor{64}\\

    \hline 
        \multirow{11}{*}{encoder} & \blockatt{24\x24}{12\x48}{48\x12}{ACT} &  \multirow{11}{*}{ B\x\wcolor{1024}\x\xycolor{32}\x\xycolor{64}}\\
        & &  \\
        & &  \\
        &MHA &  \\
        & \blockatt{24\x24}{12\x48}{48\x12}{ACT} &  \\
        & &  \\
        & &  \\
        &MHA &  \\
        & \blockatt{24\x24}{12\x48}{48\x12}{ACT} &  \\
        & &  \\
        & &  \\
         & MHA \hspace{0.7cm} MHA \hspace{0.2cm} & B\x\wcolor{1024}\x\xycolor{32}\x\xycolor{64}  \\
    \hline
        {\makecell[c]{down projection}} & \text{MLP(\wcolor{69})} \quad \text{MLP(\wcolor{69})}&  $\mu_x$: B\x\wcolor{69}\x\xycolor{32}\x\xycolor{64}\\ 
        &$\mu_x$\quad\quad\quad\quad\quad $\sigma_x$ &$\sigma_x$:B\x\wcolor{69}\x\xycolor{32}\x\xycolor{64}\\
    \hline
     latent $\y$ & $\y \leftarrow \mu_{x} + \sigma_{x} \odot \epsilon \sim \mathcal{N}(0, 1)$ & B\x \wcolor{69} \x \xycolor{32} \x \xycolor{64} \\
    \hline
        up projection & \text{MLP(\wcolor{1024})}&  B\x\wcolor{1024}\x\xycolor{\textbf{32}}\x\xycolor{64}\\ 
    \hline 
        \multirow{11}{*}{decoder} & \blockatt{24\x24}{12\x48}{48\x12}{ACT} &  \multirow{11}{*}{ B\x\wcolor{1024}\x\xycolor{32}\x\xycolor{64}}\\
        & &  \\
        & &  \\
        &MHA &  \\
        & \blockatt{24\x24}{12\x48}{48\x12}{ACT} &  \\
        & &  \\
        & &  \\
        &MHA &  \\
        & \blockatt{24\x24}{12\x48}{48\x12}{ACT} &  \\
        & &  \\
        & &  \\
        &MHA &  \\
        projection & \text{MLP(\wcolor{1104})} &   B\x\wcolor{1104}\x\xycolor{\textbf{32}}\x\xycolor{64} \\ 
        \hline
        reshape & \emph{from} \wcolor{1104} \emph{to}   \wcolor{69}\x\xycolor{4}\x\xycolor{4} &  B\x\wcolor{69}\x\xycolor{128}\x\xycolor{256} \\
	\end{tabular}
	\vspace{0.2em}
	\caption{\textbf{Architectures details of the latent transformer encoder and reconstruction transformer decoder.} The output sizes are denoted by $\{ $$\wcolor{B}$\x$\tcolor{D}$\x$\xycolor{H}$\x$\xycolor{W}$$\}$ for channel, distribution, width and height, respectively. ``CHWA($Win\_h\times Win\_w$)" indicates that wind attention is implemented within a window with shape of $h\times  w$. ``MHA" denotes the multi-head self-attention mechanism as in ViT~\cite{dosovitskiy2020image}.}
	\label{tab:arch1}
\end{table}

\begin{table}[h]
    \scriptsize
    \centering
    \tablestyle{1pt}{1.08}
    \begin{tabular}{c|c|c}
    \textbf{Stage} & \textbf{\makecell[c]{Hyperprior Encoder-Decoder}} & \textbf{Output Sizes}  \\  
    \shline
    input data $\y$ & B \x \wcolor{69}\x \xycolor{32}\x \xycolor{64} &  \\ 
       
    \hline
    patch embed & Conv \wcolor{360}\x\wcolor{69}\x\xycolor{4}\x\xycolor{4} & \wcolor{360}\x\xycolor{8}\x\xycolor{16}\\
    \hline 
 \multirow{11}{*}{encoder} & \blockatt{24\x24}{12\x48}{48\x12}{ACT} &  \multirow{11}{*}{ B\x\wcolor{360}\x\xycolor{8}\x\xycolor{16}}\\
        & &  \\
        & &  \\
        &MHA &  \\
        & \blockatt{24\x24}{12\x48}{48\x12}{ACT} &  \\
        & &  \\
        & &  \\
        &MHA &  \\
        & \blockatt{24\x24}{12\x48}{48\x12}{ACT} &  \\
        & &  \\
        & &  \\
        &MHA &  \\

    \hline
       \makecell[c]{down projection\\ hyperprior latent $\z$} & \text{MLP(\wcolor{69})}   &  B\x\wcolor{69}\x\xycolor{8}\x\xycolor{16}\\ 

    \hline
        \makecell[c]{up projection} & \text{MLP(\wcolor{360})}&  B\x\wcolor{360}\x\xycolor{\textbf{8}}\x\xycolor{16}\\ 

    \hline 
        \multirow{11}{*}{decoder} & \blockatt{24\x24}{12\x48}{48\x12}{ACT} &  \multirow{11}{*}{ B\x\wcolor{360}\x\xycolor{8}\x\xycolor{16}}\\
        & &  \\
        & &  \\
        &MHA &  \\
        & \blockatt{24\x24}{12\x48}{48\x12}{ACT} &  \\
        & &  \\
        & &  \\
        &MHA &  \\
        & \blockatt{24\x24}{12\x48}{48\x12}{ACT} &  \\
        & &  \\
        & &  \\
        &MHA &  \\

    \hline
     {\makecell[c]{ projection}} & \text{MLP(\wcolor{1104})} \quad \text{MLP(\wcolor{1104})}&  $\mu$: B\x\wcolor{1104}\x\xycolor{8}\x\xycolor{16}\\ 
        &$\mu$\quad\quad\quad\quad\quad $\sigma$ &$\sigma$:B\x\wcolor{1104}\x\xycolor{8}\x\xycolor{16}\\
    \hline
        \multirow{2}{*}{reshape} &  \multirow{2}{*}{\emph{from} \wcolor{1104}  \emph{to}   B\x\wcolor{69}\x\xycolor{4}\x\xycolor{4}} &  $\mu$: B\x\wcolor{69}\x\xycolor{32}\x\xycolor{64} \\
        &&$\sigma$ B\x\wcolor{69}\x\xycolor{32}\x\xycolor{64}
	\end{tabular}
	\vspace{0.2em}
	\caption{\textbf{Architectures details of the hyperprior transformer encoder and decoder.} The output sizes are denoted by $\{ $$\wcolor{B}$\x$\tcolor{D}$\x$\xycolor{H}$\x$\xycolor{W}$$\}$ for channel, distribution, width and height, respectively.  ``CHWA($Win\_h\times Win\_w$)" indicates that wind attention is implemented within a window with shape of $h\times w$. `MHA" denotes the multi-head self-attention mechanism as in ViT~\cite{dosovitskiy2020image}.}
	\label{tab:arch2}

\end{table}

\section{More Experiments and Analysis}
\label{sec:extra_exp}

\subsection{Evaluation Metric for Reconstruction}
To evaluate the reconstruction performance of compressed atmospheric data, we calculate the latitude-weighted Root Mean Square Error (RMSE) implemented in FengWu-GHR~\cite{han2024fengwu} as the evaluation protocols, where RMSE is a statistical metric widely used in geospatial analysis and climate science to assess the accuracy of a model's predictions or estimates of temperature, precipitation, or other meteorological variables across different latitudes. Given the prediction result $\hat{x}_{c, w, h}$ and its target (ground truth) $x_{c, w, h}$, a RMSE value is computed as follows:

\begin{equation}
\small
\label{eq:MSE}
\operatorname{RMSE}(c) =\frac{1}{T} \sum_{i=1}^{T} \sqrt{\frac{1}{W\cdot H}\sum_{w=1}^W\sum_{h=1}^H W \cdot \frac{\operatorname{cos}(\alpha_{w,h})}{\sum_{w'=1}^W \operatorname{cos}(\alpha_{w',h})}(x_{c,w,h} - \hat{x}_{c,w,h})^{2}},
\end{equation}
where $c$ denotes the index for channels that are either the surface variable or the atmosphere variable at a certain pressure level.  
$w$ and $h$ respectively denote the indices for each grid along the latitude and longitude indices. $\alpha_{w,h}$ is the latitude of point $(w,h)$. $T$ is the total number of valid test time slots.

Table~\ref{tab:wmse_compress1}, \ref{tab:wmse_compress2} and \ref{tab:wmse_compress3} respectively present the RMSE of comparing the compressed data (\textit{i.e.,} CRA5) and original data (\textit{i.e.,} ERA5) with unnormalized RMSE, which allows the comparison of atmospheric variables in their original data ranges.

\begin{table}[htbp]
\centering
\caption{The unnormlized error (error in the original data range) of CRA5 compared with ERA5 on test set (2022 year).}
\label{tab:wmse_compress1}
\begin{tabular}{|c|c|c|}
\hline
\textbf{Channel$\rightarrow$Full Name} & \textbf{Shortname} & RMSE \\
\hline
Geopotential at 1000hPa & z\_1000 & 10.427 \\
Geopotential at 950Pa & z\_950 & 8.213 \\
Geopotential at 925hPa & z\_925 & 7.583 \\
Geopotential at 900hPa & z\_900 & 7.345 \\
Geopotential at 850hPa & z\_850 & 7.490 \\
Geopotential at 800hPa & z\_800 & 7.780 \\
Geopotential at 700hPa & z\_700 & 9.167 \\
Geopotential at 600hPa & z\_600 & 9.889 \\
Geopotential at 500hPa & z\_500 & 11.682 \\
Geopotential at 400hPa & z\_400 & 14.469 \\
Geopotential at 300hPa & z\_300 & 13.071 \\
Geopotential at 250hPa & z\_250 & 13.028 \\
Geopotential at 200hPa & z\_200 & 15.829 \\
Geopotential at 150hPa & z\_150 & 17.708 \\
Geopotential at 100hPa & z\_100 & 14.977 \\
Geopotential at 70hPa & z\_70 & 19.613 \\
Geopotential at 50hPa & z\_50 & 21.136 \\
Geopotential at 30hPa & z\_30 & 21.568 \\
Geopotential at 20hPa & z\_20 & 25.908 \\
Geopotential at 10hPa & z\_10 & 23.015 \\
Geopotential at 7hPa & z\_7 & 22.738 \\
Geopotential at 5hPa & z\_5 & 25.798 \\
Geopotential at 3hPa & z\_3 & 36.266 \\
Geopotential at 2hPa & z\_2 & 37.017 \\
Geopotential at 1hPa & z\_1 & 44.318 \\
Specific humidity at 1000hPa& q\_1000 & 0.00035 \\
Specific humidity at 950hPa & q\_950 & 0.00037 \\
Specific humidity at 925hPa & q\_925 & 0.00039 \\
Specific humidity at 1000hPa & q\_900 & 0.00041 \\
Specific humidity at 850hPa & q\_850 & 0.00044 \\
Specific humidity at 800hPa & q\_800 & 0.00042 \\
Specific humidity at 700hPa & q\_700 & 0.00031 \\
Specific humidity at 600hPa & q\_600 & 0.00022 \\
Specific humidity at 500hPa & q\_500 & 0.00014 \\
Specific humidity at 400hPa & q\_400 & 0.00007 \\
Specific humidity at 300hPa & q\_300 & 0.00002 \\
Specific humidity at 250hPa & q\_250 & 0.00001 \\
Specific humidity at 200hPa & q\_200 & 0.00000 \\
Specific humidity at 150hPa & q\_150 & 0.00000 \\
Specific humidity at 100hPa & q\_100 & 0.00000 \\
Specific humidity at 70hPa & q\_70 & 0.00000 \\
Specific humidity at 50hPa & q\_50 & 0.00000 \\
Specific humidity at 30hPa & q\_30 & 0.00000 \\
Specific humidity at 20hPa & q\_20 & 0.00000 \\
Specific humidity at 10hPa & q\_10 & 0.00000 \\
Specific humidity at 7hPa & q\_7 & 0.00000 \\
Specific humidity at 5hPa & q\_5 & 0.00000 \\
Specific humidity at 3hPa & q\_3 & 0.00000 \\
Specific humidity at 2hPa & q\_2 & 0.00000 \\
Specific humidity at 1hPa & q\_1 & 0.00000 \\
\bottomrule
    \end{tabular}
    \label{tab:my_label}
\end{table}

\begin{table}[htbp]
\centering
\caption{The unnormlized error (error in the original data range) of CRA5 compared with ERA5 on test set (2022 year).}
\label{tab:wmse_compress2}
\begin{tabular}{|c|c|c|}
\hline
\textbf{Channel$\rightarrow$Full Name} & \textbf{Shortname} & RMSE \\
\hline
Longitudinal wind speed at 1000hPa & u\_1000 & 0.346 \\
Longitudinal wind speed at 950hPa & u\_950 & 0.427 \\
Longitudinal wind speed at 925hPa & u\_925 & 0.431 \\
Longitudinal wind speed at 900hPa & u\_900 & 0.478 \\
Longitudinal wind speed at 850hPa & u\_850 & 0.567 \\
Longitudinal wind speed at 800hPa & u\_800  & 0.582  \\
Longitudinal wind speed at 700hPa & u\_700 & 0.568 \\
Longitudinal wind speed at 600hPa & u\_600 & 0.537 \\
Longitudinal wind speed at 500hPa & u\_500 & 0.536 \\
Longitudinal wind speed at 400hPa & u\_400 & 0.557 \\
Longitudinal wind speed at 300hPa & u\_300 & 0.582 \\
Longitudinal wind speed at 250hPa & u\_250 & 0.575 \\
Longitudinal wind speed at 200hPa & u\_200 & 0.535 \\
Longitudinal wind speed at 150hPa & u\_150 & 0.450 \\
Longitudinal wind speed at 100hPa & u\_100 & 0.337 \\
Longitudinal wind speed at 70hPa & u\_70 & 0.290 \\
Longitudinal wind speed at 50hPa & u\_50 & 0.276 \\
Longitudinal wind speed at 30hPa & u\_30 & 0.271 \\
Longitudinal wind speed at 20hPa & u\_20 & 0.284 \\
Longitudinal wind speed at 10hPa & u\_10 & 0.278 \\
Longitudinal wind speed at 7hPa & u\_7 & 0.261 \\
Longitudinal wind speed at 5hPa & u\_5 & 0.255 \\
Longitudinal wind speed at 3hPa & u\_3 & 0.294 \\
Longitudinal wind speed at 2hPa& u\_2 & 0.307 \\
Longitudinal wind speed at 1hPa& u\_1 & 0.333 \\
Meridional wind speed at 1hPa & v\_1000 & 0.336 \\
Meridional wind speed at 1hPa & v\_950 & 0.413 \\
Meridional wind speed at 1hPa & v\_925 & 0.412 \\
Meridional wind speed at 1hPa & v\_900 & 0.461 \\
Meridional wind speed at 1hPa & v\_850 & 0.543 \\
Meridional wind speed at 1hPa & v\_800 & 0.557 \\
Meridional wind speed at 1hPa & v\_700 & 0.532 \\
Meridional wind speed at 1hPa & v\_600 & 0.494 \\
Meridional wind speed at 1hPa & v\_500 & 0.493 \\
Meridional wind speed at 1hPa & v\_400 & 0.523 \\
Meridional wind speed at 1hPa & v\_300 & 0.544 \\
Meridional wind speed at 1hPa & v\_250 & 0.513 \\
Meridional wind speed at 1hPa & v\_200 & 0.476 \\
Meridional wind speed at 1hPa & v\_150 & 0.396 \\
Meridional wind speed at 1hPa & v\_100 & 0.280 \\
Meridional wind speed at 1hPa & v\_70 & 0.232 \\
Meridional wind speed at 1hPa & v\_50 & 0.210 \\
Meridional wind speed at 1hPa & v\_30 & 0.196 \\
Meridional wind speed at 1hPa & v\_20 & 0.201 \\
Meridional wind speed at 1hPa & v\_10 & 0.203 \\
Meridional wind speed at 1hPa & v\_7 & 0.190 \\
Meridional wind speed at 1hPa & v\_5 & 0.183 \\
Meridional wind speed at 1hPa & v\_3 & 0.218 \\
Meridional wind speed at 1hPa & v\_2 & 0.219 \\
Meridional wind speed at 1hPa & v\_1 & 0.225 \\
\bottomrule
\end{tabular}
\label{tab:my_label}
\end{table}

\begin{table}[htbp]
\centering
\caption{The unnormlized error (error in the original data range) of CRA5 compared with ERA5 on test set (2022 year).}
\label{tab:wmse_compress3}
\begin{tabular}{|c|c|c|}
\hline
\textbf{Channel$\rightarrow$Full Name} & \textbf{Shortname} & RMSE \\
\hline
Atmospheric temperature at 1 hPa & t\_1000 & 0.417 \\
Atmospheric temperature at 1 hPa & t\_950 & 0.365 \\
Atmospheric temperature at 1 hPa & t\_925 & 0.343 \\
Atmospheric temperature at 1 hPa & t\_900 & 0.344 \\
Atmospheric temperature at 1 hPa & t\_850 & 0.337 \\
Atmospheric temperature at 1 hPa & t\_800 & 0.309 \\
Atmospheric temperature at 1 hPa & t\_700 & 0.260 \\
Atmospheric temperature at 1 hPa & t\_600 & 0.232 \\
Atmospheric temperature at 1 hPa & t\_500 & 0.203 \\
Atmospheric temperature at 1 hPa & t\_400 & 0.194 \\
Atmospheric temperature at 1 hPa & t\_300 & 0.169 \\
Atmospheric temperature at 1 hPa & t\_250 & 0.166 \\
Atmospheric temperature at 1 hPa & t\_200 & 0.162 \\
Atmospheric temperature at 1 hPa & t\_150 & 0.160 \\
Atmospheric temperature at 1 hPa & t\_100 & 0.180 \\
Atmospheric temperature at 1 hPa & t\_70 & 0.172 \\
Atmospheric temperature at 1 hPa & t\_50 & 0.152 \\
Atmospheric temperature at 1 hPa & t\_30 & 0.141 \\
Atmospheric temperature at 1 hPa & t\_20 & 0.149 \\
Atmospheric temperature at 1 hPa & t\_10 & 0.151 \\
Atmospheric temperature at 1 hPa & t\_7 & 0.148 \\
Atmospheric temperature at 1 hPa & t\_5 & 0.143 \\
Atmospheric temperature at 1 hPa & t\_3 & 0.170 \\
Atmospheric temperature at 1 hPa & t\_2 & 0.161 \\
Atmospheric temperature at 1 hPa & t\_1 & 0.150 \\

Vertical velocity speed at 1000 hPa & w\_1000 & 0.051 \\
Vertical velocity speed at 950 hPa & w\_950 & 0.069 \\
Vertical velocity speed at 925 hPa & w\_925 & 0.076 \\
Vertical velocity speed at 900 hPa & w\_900 & 0.084 \\
Vertical velocity speed at 850 hPa & w\_850 & 0.102 \\
Vertical velocity speed at 800 hPa & w\_800 & 0.118 \\
Vertical velocity speed at 700 hPa & w\_700 & 0.134 \\
Vertical velocity speed at 600 hPa & w\_600 & 0.129 \\
Vertical velocity speed at 500 hPa & w\_500 & 0.117 \\
Vertical velocity speed at 400 hPa & w\_400 & 0.107 \\
Vertical velocity speed at 300 hPa & w\_300 & 0.075 \\
Vertical velocity speed at 250 hPa & w\_250 & 0.061 \\
Vertical velocity speed at 200 hPa & w\_200 & 0.050 \\
Vertical velocity speed at 150 hPa & w\_150 & 0.035 \\
Vertical velocity speed at 100 hPa & w\_100 & 0.013 \\
Vertical velocity speed at 70 hPa & w\_70 & 0.005 \\
Vertical velocity speed at 50 hPa & w\_50 & 0.003 \\
Vertical velocity speed at 30 hPa & w\_30 & 0.001 \\
Vertical velocity speed at 20 hPa & w\_20 & 0.001 \\
Vertical velocity speed at 10 hPa & w\_10 & 0.001 \\
Vertical velocity speed at 7 hPa & w\_7 & 0.000 \\
Vertical velocity speed at 5 hPa & w\_5 & 0.000 \\
Vertical velocity speed at 3 hPa & w\_3 & 0.000 \\
Vertical velocity speed at 2 hPa & w\_2 & 0.000 \\
Vertical velocity speed at 1 hPa & w\_1 & 0.000 \\
10m v-component of wind & 10v & 0.298 \\
10m u-component of wind & 10u & 0.308 \\
100m v-component of wind & 100v & 0.352 \\
100m u-component of wind & 100u & 0.364 \\
2m temperature & t2m & 0.664 \\
Total cloud cover & tcc & 0.121 \\
surface pressure & sp & 376.906 \\
Total precipitation (6hour) & tp6h & 0.800 \\
Mean sea-level pressure & msl & 13.551 \\
\bottomrule
\end{tabular}
\label{tab:my_label}
\end{table}

\subsection{Comparison with Traditional Image Codecs}
Apart from comparing with the existing neural image compression methods~\cite{minnen2018joint,balle2018variational,cheng2020learned,xie2021enhanced,he2022elic,liu2023learned}, we also attempt the traditional image codec JPEG2000~\cite{taubman2002jpeg2000} to compress the atmospheric data. It adopts Discrete Wavelet Transform (DWT) and still serves as one of the most commonly used standards for compressing the image data. Here, we attempt to compress the weather data with JPEG2000, where we use OpenJPEG~\footnote{\url{https://github.com/uclouvain/openjpeg}} part-1 to compress each individual slice into a compressed j2k file. Table~\ref{tab:comparison_jpeg} reports the compression results of JPEG2000 on our test set, ERA5: $128\times256$, 2022 year (please refer to Sec. 4.1 in the manuscript for more details).

\begin{figure}
  \centering      
    \includegraphics[width=0.99\linewidth]{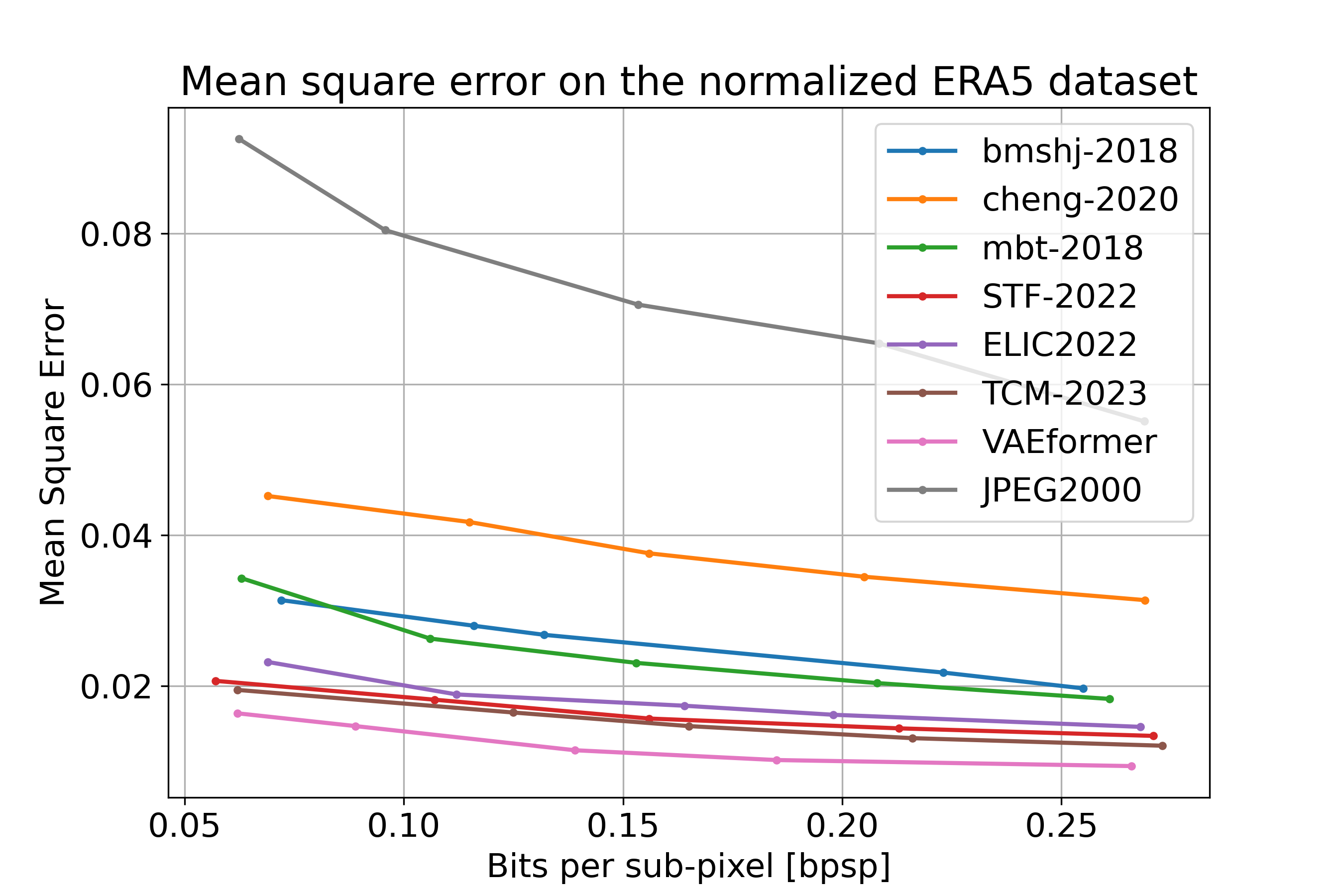}
    \caption{
The Rate-Distortion (RD) performance on test data, ``ERA5: 128$\times$256, 2022 year", between those NIC methods (bmsj2018, cheng2020,  mbt2018, STF2022,  ELIC2022, TCM2023 and VAEformer (ours)) and traditional image codec JPEG2000. Here, the degree of distortion is measured based on the mean squared error (MSE).}
    \label{fig:RD_Curve_supp}
\end{figure}

The Fig.~\ref{fig:RD_Curve_supp} demonstrates that JPEG2000 yields significantly higher MSE than all other NIC methods. This finding suggests that traditional image compression techniques such as JPEG2000 may not be as effective as learning-based methods for compressing atmospheric data. We intend to carry out additional research to evaluate the performance of both traditional and neural image codecs in compressing atmospheric data. We believe that these studies will provide valuable insights and benefits to the weather research community in the future.

\begin{table}[t]
	\centering
	\caption{The compression performance on the test set (12.3GB) with JPEG2000~\cite{taubman2002jpeg2000}.}
	\label{tab:comparison_jpeg}

	\resizebox{1.0\linewidth}{!}{
	\tablestyle{1.5pt}{1.0}
	\begin{tabular}{@{}lcccc@{}}
		\toprule
		Method & File Size (MB)&\makecell[c]{Overall MSE $\times$100}&\makecell[c]{Compression Ratio} &\makecell[c]{bpsp}\\
        JPEG2000-bit16-dB30&82   &6.44     &153& 0.208,
      
       \\
	JPEG2000-bit16-dB32&105  &4.00  &119 &0.269 \\
        JPEG2000-bit16-dB34&137  &2.46  &92  &0.348\\
        JPEG2000-bit16-dB36&182  &1.56  &64  &0.463\\
        JPEG2000-bit16-dB38&236  &0.87  &53  &0.600   \\
        JPEG2000-bit16-dB40&304  &\textbf{0.54} &41  &0.773  \\
         VAEformer ($\lambda$=1e-3)  &\textbf{61} &1.14& \textbf{206} & \textbf{0.155}	\\
         \bottomrule
	\end{tabular}}
\end{table}

\begin{figure*}[!t]
  \centering      
    \includegraphics[width=0.85\textwidth]{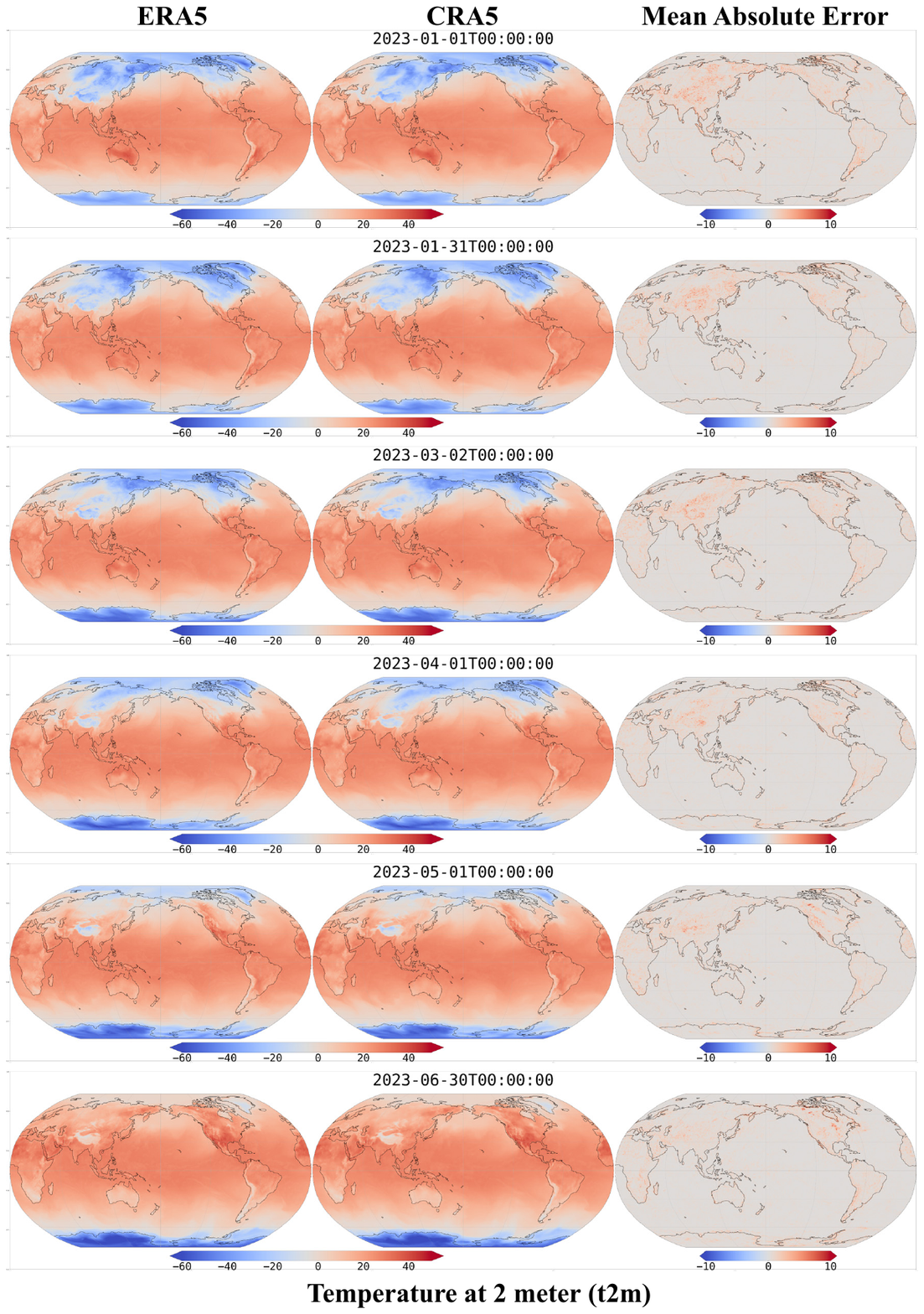}
    \caption{Visualization samples of t2m on the ERA5 and th ecompressed CRA5. From the left to the right column: ERA5, CRA5, and their mean absolute error map. }
    \label{fig:supp_diff_t2m}
\end{figure*}

\begin{figure*}[!t]
  \centering      \includegraphics[width=0.85\textwidth]{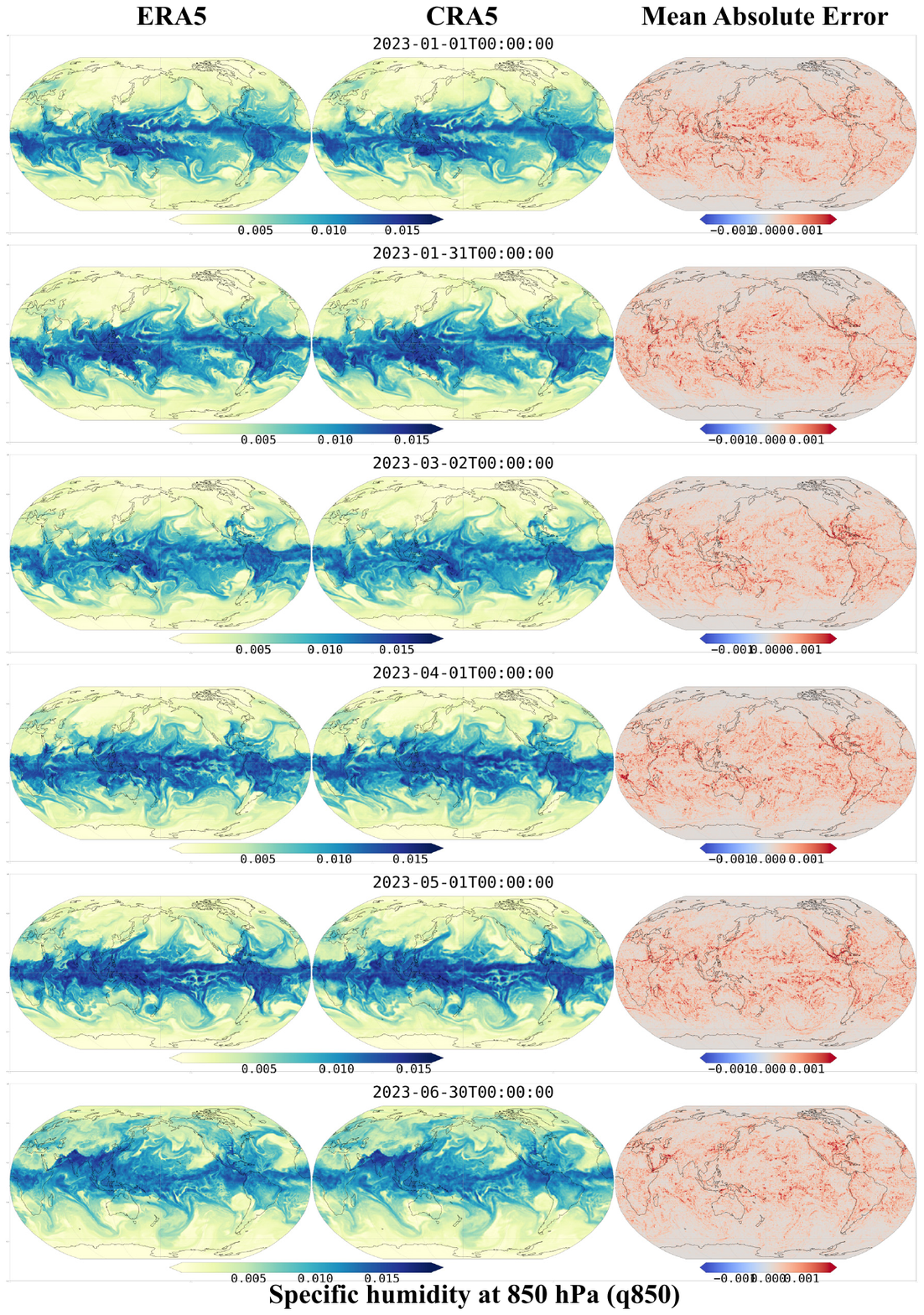}
    \caption{Visualization samples of t2m on the ERA5 and th ecompressed CRA5. From the left to the right column: ERA5, CRA5, and their mean absolute error map. }
    \label{fig:supp_diff_q850}
\end{figure*}
\begin{figure*}[!t]
  \centering      \includegraphics[width=0.85\textwidth]{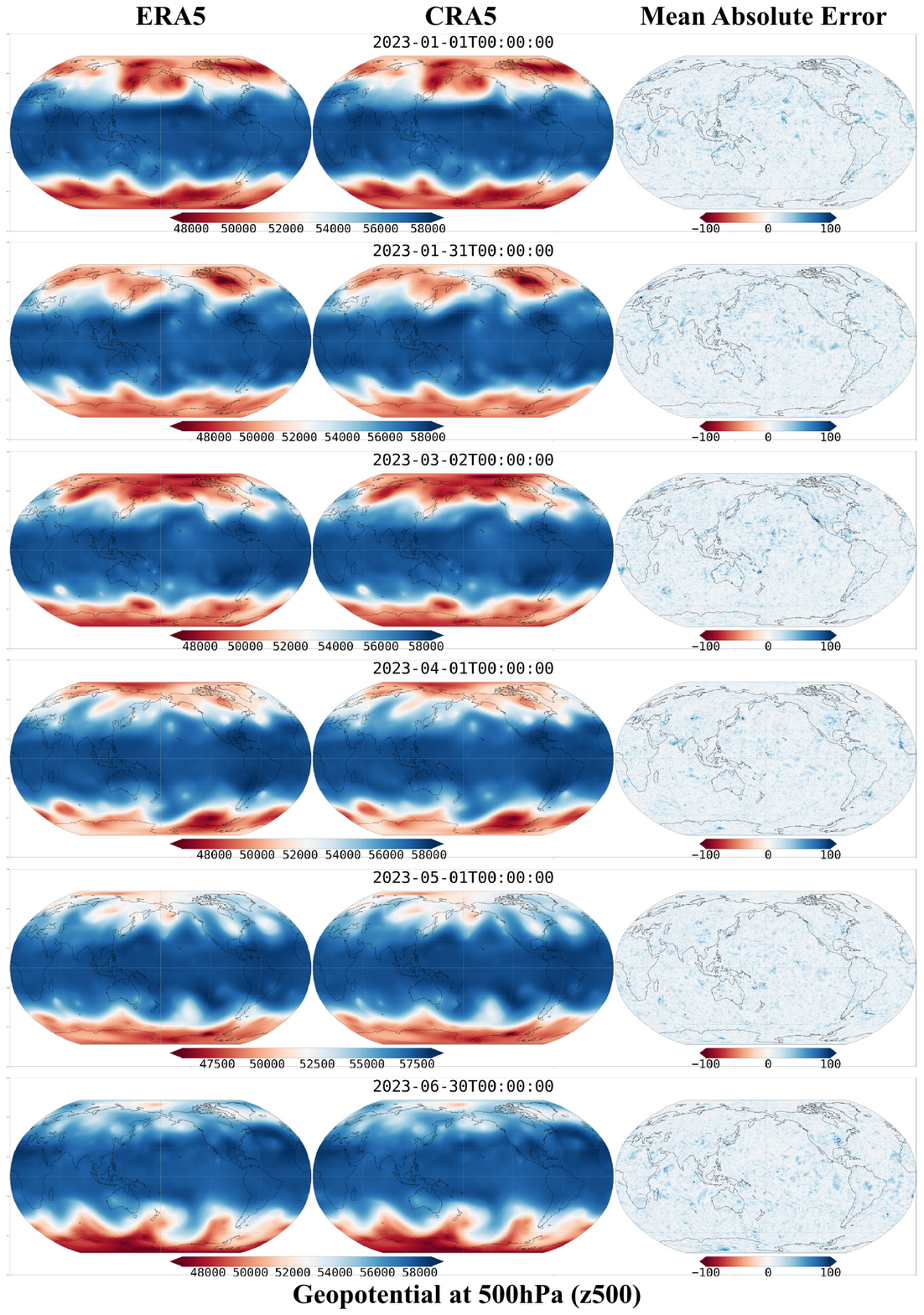}
    \caption{Visualization samples of z500 on the ERA5 and th ecompressed CRA5. From the left to the right column: ERA5, CRA5, and their mean absolute error map.  }
    \label{fig:supp_diff_z500}
\end{figure*}
\begin{figure*}[!t]
  \centering      \includegraphics[width=0.85\textwidth]{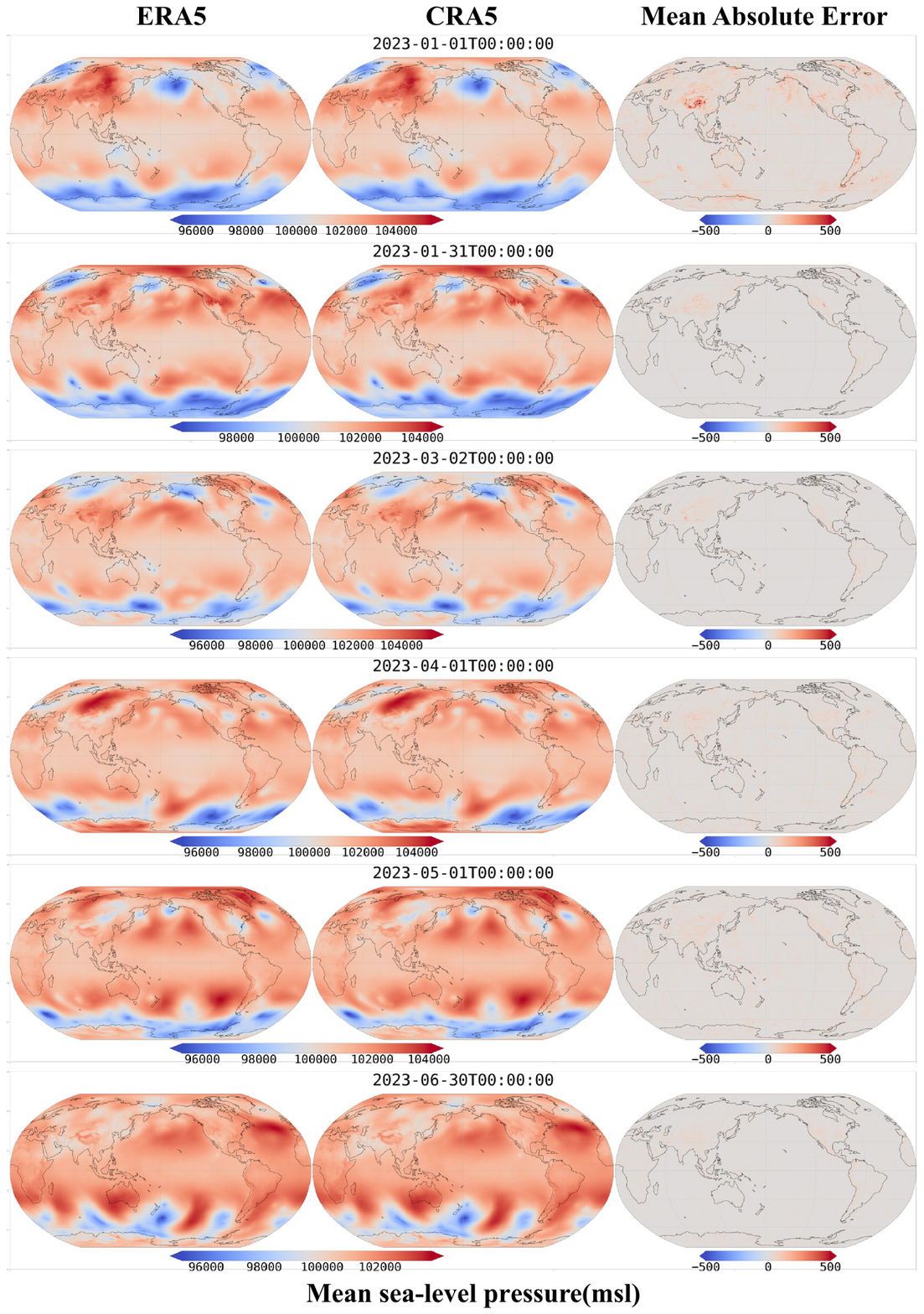}
    \caption{Visualization samples of msl on the ERA5 and th ecompressed CRA5. From the left to the right column: ERA5, CRA5, and their mean absolute error map. }
    \label{fig:supp_diff_msl}
\end{figure*}

\begin{figure*}[!t]
  \centering      \includegraphics[width=0.85\textwidth]{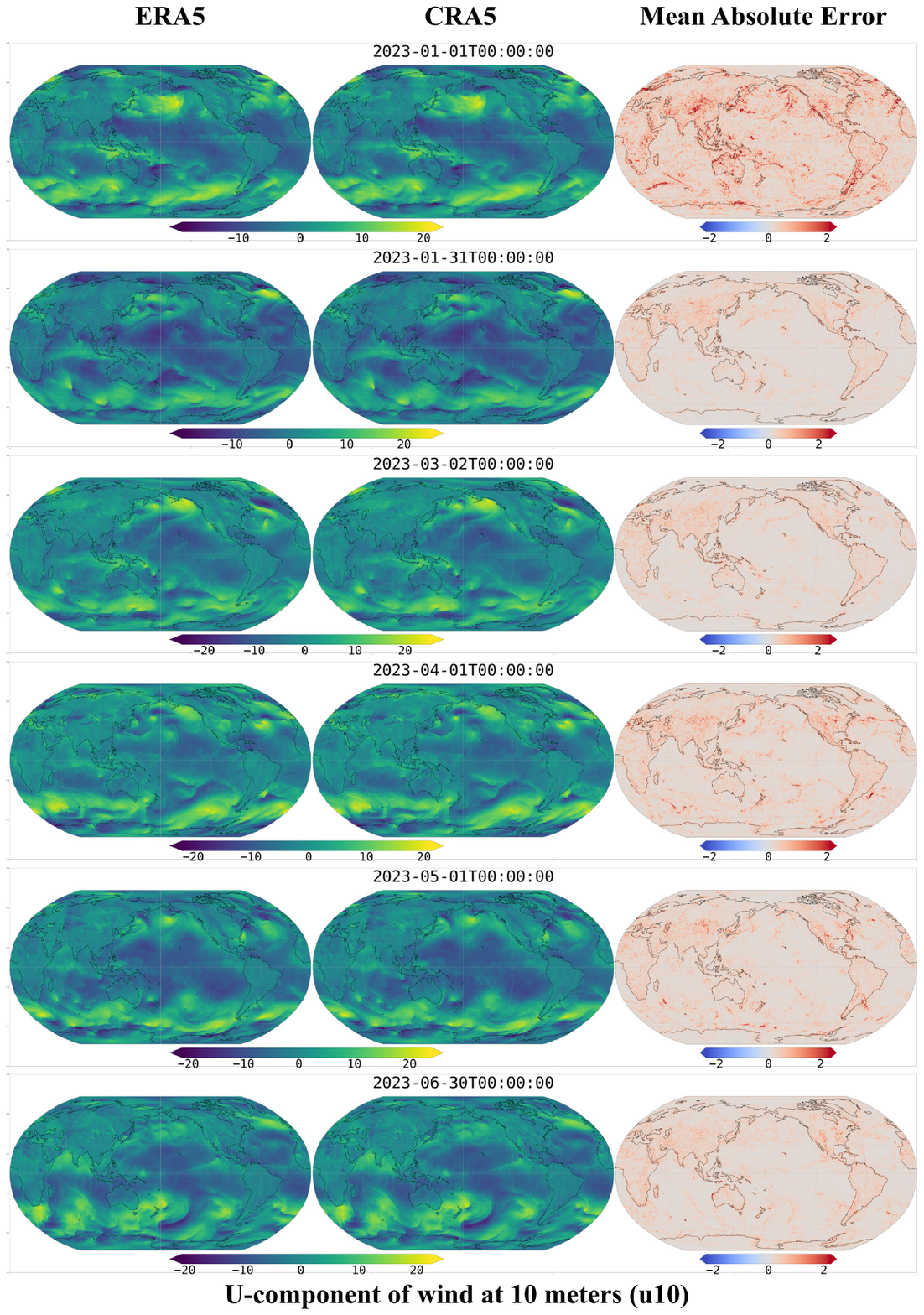}
    \caption{Visualization samples of u10 on the ERA5 and th ecompressed CRA5. From the left to the right column: ERA5, CRA5, and their mean absolute error map.  }
    \label{fig:supp_diff_u10}
\end{figure*}

\begin{figure*}[!t]
  \centering      \includegraphics[width=0.95\textwidth]{figs/supp_vis1.pdf}
    \caption{Visualization maps of the reconstructed weather data. From the left to the right column: ERA5, VAEformer-69, and Cheng2020-attn~\cite{cheng2020learned}. }
    \label{fig:vis_supp1}
\end{figure*}

\begin{figure*}[!t]
  \centering      \includegraphics[width=0.95\textwidth]{figs/supp_vis2.pdf}
    \caption{Visualization maps of the reconstructed weather data. From the left to the right column: ERA5, VAEformer-69, and Cheng2020-attn~\cite{cheng2020learned}. }
    \label{fig:vis_supp2}
\end{figure*}

\section{Visualization of Reconstructed Samples}
\label{sec:vis}

We visualize some variables in the compressed CAR5 dataset and the ERA5 dataset in Figure~\ref{fig:supp_diff_t2m}, Figure~\ref{fig:supp_diff_q850}, Figure~\ref{fig:supp_diff_z500}, Figure~\ref{fig:supp_diff_msl}, and Figure~\ref{fig:supp_diff_u10}, where the Mean Absolute Error map shows their difference in different regions. Figure~\ref{fig:vis_supp1} and Figure~\ref{fig:vis_supp2} compare the qualitative performance of the VAEformer with a representative NIC method, Cheng2020-attn~\cite{cheng2020learned}. Those samples are chosen from the same day at four timestamps (\emph{i.e.,} T00:00:00, T06:00:00,  T12:00:00, and  T18:00:00) in the test set. We can see that even under the same compression ratio (VAEformer $\lambda=10$ \emph{v.s} Cheng2020-attn),
VAEformer can produce better perceptual reconstructed results. These examples suggest that our VAEformer is capable of learning more representative features that effectively capture the detailed information in the atmospheric data.

\section{Limitations}
\label{sec:limit}
Although we have significantly reduced the cost of storing and transmitting atmospheric data, the practical application of this technology still requires clients to possess substantial computing resources to operate the decompression model, typically involving GPUs. For end users with limited GPU capabilities, decompressing data can be time-consuming. Our goal is to further reduce the computational demands without compromising compression efficiency, potentially through the use of advanced model compression techniques (\textit{e.g.,} channel pruning).

\section{Potential negative societal impact}
\label{sec:potential}
The training of the model involved in this technology requires significant computational resources, which consume a considerable amount of energy. This energy consumption has an environmental impact, as it contributes to carbon emissions and other pollutants. Therefore, we recommend the use of clean energy sources, such as renewable energy, to decrease the environmental impact of this technology.

By using clean energy sources, we can significantly reduce the carbon footprint of the model training process and contribute to a more sustainable future. It is essential to consider the environmental impact of new technologies and take proactive measures to mitigate their negative effects. This can be achieved through the use of energy-efficient hardware, the optimization of algorithms to reduce computational requirements, and the use of clean energy sources. By promoting sustainable practices, we can ensure that the development and implementation of new technologies do not come at the cost of our environment.



\bibliographystyle{ACM-Reference-Format}
\bibliography{sample-base}